\documentclass[preprint,12pt]{elsarticle}

\usepackage{amssymb}
\usepackage{amsmath}
\usepackage{graphicx}
\usepackage{xcolor}
\graphicspath{{./}}
\usepackage{booktabs}
\usepackage{multirow}
\usepackage{subcaption}
\usepackage{hyperref}
\usepackage{algorithm}
\usepackage{algorithmic}

\journal{Energy}

\begin{document}

\begin{frontmatter}

\title{ECTO: Exogenous-Conditioned Temporal Operator for Ultra-Short-Term Wind Power Forecasting}

\author[addr1]{Liyaqin Li\fnref{equal}}
\author[addr1]{Junjun Wang\fnref{equal}}
\author[addr1]{Jianxiang Li}
\author[addr1]{Wei Huang}
\author[addr2]{Qianhui Yu}
\author[addr1]{Cao Yuan\corref{cor1}}
\ead{yc@whpu.edu.cn}
\cortext[cor1]{Corresponding author.}
\fntext[equal]{These authors contributed equally to this work.}

\affiliation[addr1]{organization={Wuhan Polytechnic University},
            city={Wuhan},
            postcode={430048},
            state={Hubei},
            country={China}}
\affiliation[addr2]{organization={Wuhan Public Meteorological Service Center},
            city={Wuhan},
            postcode={430040},
            state={Hubei},
            country={China}}

\begin{abstract}
\sloppy
Accurate ultra-short-term wind power forecasting is critical for grid dispatch and reserve management, yet remains challenging due to the non-stationary, condition-dependent nature of wind generation. Meteorological exogenous variables carry substantial predictive information, but the most informative variable combination---and the way these variables should influence the forecast---varies across sites, operating conditions, and prediction horizons. Existing deep learning approaches either treat exogenous inputs as generic auxiliary channels through uniform mixing or soft gating, or rely on fixed preprocessing steps such as PCA, without exploiting the physical structure of meteorological variables or adapting to regime-dependent predictive relationships. We propose ECTO (Exogenous-Conditioned Temporal Operator), a unified framework that decomposes exogenous variable modeling into two complementary modules. Physically-Grounded Variable Selection (PGVS) performs hierarchical, group-aware sparse selection over exogenous variables using a domain-informed physical prior and sparsemax activations, producing a compact, condition-adaptive exogenous context. Exogenous-Conditioned Regime Refinement (ECRR) routes the forecast through a small number of learned regime experts that apply gain--bias calibration and horizon-specific corrections, following a mixture-of-experts paradigm. Experiments on three wind farms spanning different climates, capacities (66--200~MW), and exogenous dimensions (11--13 variables) demonstrate that ECTO achieves the lowest MSE across all sites, with relative improvements over the strongest baseline ranging from 2.2\% to 5.2\%, widening to 8.6\% at the longer prediction horizon ($H{=}32$). Ablation analysis confirms that each exogenous-related component contributes positively (PGVS +1.84\%, ECRR +2.86\%), and interpretability analysis reveals that PGVS learns physically meaningful, site-specific variable selection patterns, while ECRR converges to well-separated calibration strategies consistent across sites.
\end{abstract}


\end{frontmatter}

\section{Introduction}
\label{sec:intro}
\sloppy

Wind power has become a cornerstone of the global energy transition, yet its integration into power systems remains challenging due to the inherent intermittency, volatility, and non-stationarity of wind generation~\citep{tawn2022review,wang2021review}. Unlike load or solar photovoltaic output, wind power time series exhibit weak periodicity and statistical properties that change drastically with meteorological conditions~\citep{liu2019data}. Among the most demanding forecasting scenarios are \emph{ramp events}---sudden, large-magnitude power swings driven by cold-front passages, low-pressure systems, or low-level jets---which can exceed tens of megawatts within minutes and pose acute risks to grid dispatch and reserve management~\citep{gallego2015review}. Such events can trigger frequency deviations, require costly fast-responding reserves, and, in severe cases, lead to forced curtailment~\citep{gallego2015identifying}.

These characteristics originate in the atmosphere: wind power generation is governed by complex atmospheric processes in which wind speed, temperature, pressure, humidity, and wind direction jointly determine the energy conversion efficiency~\citep{dalton2022exogenous,kirchner2024hybrid}. This physical reality implies that the \emph{most informative} set of exogenous variables---and the \emph{way} they should influence the forecast---may differ substantially across sites, operating conditions, and prediction horizons~\citep{dalton2022exogenous}. For instance, at low wind speeds the air-density correction is comparable to the wind-speed contribution, whereas at high wind speeds the cubic relationship dominates; wind direction, meanwhile, influences inter-turbine wake effects only within specific directional sectors. This condition-dependent nature of exogenous information motivates a modular architecture that can adaptively select and condition on the relevant variables for each prediction.

Despite this well-established physical motivation, most existing deep learning approaches to wind power forecasting do not treat exogenous variable modeling as a first-class design problem. Methods targeting the wind energy domain have made substantial progress in encoding temporal patterns and handling non-stationarity~\citep{wang2025frequency}, yet when exogenous meteorological variables are included, they are typically incorporated through PCA-based dimensionality reduction or direct concatenation, without structured mechanisms for selecting, grouping, or conditioning on the most informative variables. Beyond wind energy, the broader time series forecasting literature has explored exogenous variable handling---including channel-independent strategies~\citep{zeng2023transformers,nie2023patchtst}, uniform variate-token mixing~\citep{liu2024itransformer}, soft-gated fusion~\citep{wang2024timexer,chen2026xlinear}, and variable selection networks~\citep{lim2021temporal}---but these general solutions do not fully exploit the physical group structure of meteorological data or adapt to regime-dependent predictive relationships.

To address these challenges, we propose the \textbf{Exogenous-Conditioned Temporal Operator} (\textbf{ECTO}), a unified forecasting framework that moves beyond simple fusion or concatenation by decomposing exogenous variable modeling into two complementary questions: \emph{which} exogenous variables matter for the current prediction, and \emph{how} should they modulate the forecast under different operating conditions. ECTO encodes the target power sequence through a hierarchical convolutional Temporal Tokenizer and conditions the resulting forecast on a compact, adaptively selected exogenous context. The main contributions are as follows:
\begin{enumerate}
\item We propose PGVS, a physically-guided hierarchical sparse selection module that leverages domain-informed variable grouping and sparsemax activations to automatically identify the most informative meteorological variables under different operating conditions.
\item We propose ECRR, an end-to-end regime-aware mixture-of-experts refinement that dynamically routes the forecast through learned regime experts without pre-clustering, applying gain--bias calibration and horizon-specific corrections.
\item We demonstrate that ECTO achieves the lowest MSE across three wind farms with different climates and capacities, with the advantage widening from 3.7\% at $H{=}16$ to 8.6\% at $H{=}32$, and show that the learned variable selection and calibration patterns are physically interpretable and consistent across sites.
\end{enumerate}

The remainder of this paper is organized as follows. Section~\ref{sec:related} reviews related work on wind power forecasting, exogenous variable modeling, and regime-aware prediction. Section~\ref{sec:method} formulates the forecasting problem and presents the ECTO architecture in detail. Section~\ref{sec:exp} describes the experimental setup and reports main results, ablation studies, and interpretability analysis. Section~\ref{sec:conclusion} concludes with a discussion of limitations and future directions.

\section{Related Work}
\label{sec:related}

\subsection{Wind Power Forecasting}
\label{sec:rw-wpf}

Wind power forecasting methods can be broadly categorized into physical, statistical, and deep learning approaches. Physical models rely on numerical weather prediction (NWP) systems that solve the governing equations of the atmosphere to simulate future wind fields~\citep{olson2019improving, al2010review}. While effective for medium- to long-range horizons, they are computationally expensive and often exhibit systematic biases at the turbine level that require statistical post-processing~\citep{jung2014current}.

Statistical models, including autoregressive integrated moving average (ARIMA) and its variants, have been widely adopted for short-term forecasting due to their simplicity and interpretability~\citep{tawn2022review}. However, their linear formulation limits their ability to capture the strongly nonlinear and non-stationary dynamics inherent in wind power generation~\citep{liu2019data}.

The advent of deep learning has substantially advanced the state of the art. Recurrent architectures such as LSTM~\citep{hochreiter1997lstm} and GRU~\citep{cho2014gru} excel at capturing temporal dependencies~\citep{wang2021review}, while convolutional networks extract local temporal features. More recently, Transformer-based models have been explored for wind power forecasting, leveraging self-attention to model long-range temporal dependencies~\citep{wang2025frequency}. Nevertheless, as demonstrated in the broader time series forecasting literature, even carefully designed Transformer architectures do not consistently outperform simple linear baselines on several long-term time series forecasting benchmarks~\citep{zeng2023transformers}, underscoring the importance of appropriate inductive biases.

A common thread across these approaches is that the primary modeling effort is directed toward encoding the temporal structure of the target power series. Exogenous meteorological variables, when included, are typically incorporated through straightforward concatenation or uniform channel mixing, without explicit mechanisms for selecting, grouping, or conditioning on the most informative variables under varying meteorological regimes.

\subsection{Exogenous Variable Modeling}
\label{sec:rw-exo}

The treatment of exogenous variables in time series forecasting has evolved along several lines, yet most existing strategies do not fully exploit the physically motivated group structure inherent in meteorological data.

\paragraph{Channel-independent and uniform mixing}
A number of recent multivariate forecasting methods adopt a channel-independent strategy, in which each variate is modeled as a univariate series without cross-channel interaction~\citep{zeng2023transformers, nie2023patchtst}. While this avoids spurious cross-variable noise, it entirely discards exogenous information. At the other extreme, approaches such as iTransformer~\citep{liu2024itransformer} treat every variate as an independent token and apply inverted attention across variates, without distinguishing the asymmetric roles of the target and exogenous channels.

\paragraph{Exogenous-aware fusion}
Several methods have been proposed to explicitly incorporate exogenous variables. TimeXer~\citep{wang2024timexer} introduces exogenous tokens that interact with endogenous representations through cross-attention, but does not perform variable selection---all exogenous channels contribute equally. XLinear~\citep{chen2026xlinear} employs a Sigmoid-gated block to control exogenous influence, yet the continuous soft gating ensures that every variable retains a non-zero contribution, potentially allowing irrelevant variables to dilute the forecast signal. CrossLinear~\citep{zhou2025crosslinear} embeds cross-correlation between exogenous and target channels via plug-and-play modules, but similarly lacks a mechanism for structured variable selection or regime-dependent modulation.

\paragraph{Variable selection}
Temporal Fusion Transformers (TFT)~\citep{lim2021temporal} incorporate a variable selection network that generates instance-wise feature weights. While this enables adaptive input relevance scoring, the selection operates at the level of individual features rather than physically coherent groups, and is not specifically designed for the domain structure of meteorological variables. ExoTST~\citep{tayal2024exotst} (arXiv preprint) and GCGNet~\citep{li2026gcgnet} further explore exogenous-aware architectures, though both assume access to future exogenous observations---a setting that is not applicable to ultra-short-term wind power forecasting, where only historical exogenous data are available at prediction time.

\paragraph{Domain-specific exogenous treatment}
Within the wind energy literature, the value of meteorological exogenous variables for wind speed and power prediction has been extensively documented~\citep{dalton2022exogenous, kirchner2024hybrid}. However, the optimal variable combination depends on the site, atmospheric height, and prediction horizon, and no universally best subset has been identified~\citep{dalton2022exogenous}. Wanek~\citep{wanek2026variable} demonstrated that even simple multilayer perceptrons can achieve competitive forecasting performance when paired with systematically processed exogenous inputs. This result highlights that \emph{how} exogenous variables are selected and organized may matter more than the complexity of the downstream model. Nonetheless, most wind-specific deep learning models either rely on PCA-based dimensionality reduction~\citep{wang2025frequency} or direct concatenation, without leveraging the physical group structure of meteorological variables.

\paragraph{ECTO's position}
ECTO departs from the above paradigms by introducing a \emph{Physically-Grounded Variable Selection} (PGVS) module that performs hierarchical, group-aware hard selection over exogenous variables using sparsemax activations~\citep{martins2016softmax} (which can produce exact zeros), and an \emph{Exogenous-Conditioned Regime Refinement} (ECRR) module that dynamically adjusts the forecast based on the selected exogenous context. Together, these modules address both \emph{which} exogenous variables matter and \emph{how} they should modulate the prediction under different operating conditions.

\subsection{Regime-Aware Prediction}
\label{sec:rw-regime}

The idea that predictive relationships in wind power generation change across meteorological regimes has a long history in the forecasting literature.

\paragraph{Statistical regime-switching}
Gneiting et al.\ \citep{gneiting2006calibrated} proposed a regime-switching space-time method for calibrated probabilistic wind forecasting, in which distinct predictive models are fitted for predefined weather regimes based on wind direction and atmospheric stability. Browell et al.~\citep{browell2018improved} showed that identifying atmospheric regimes can improve very-short-term spatio-temporal wind forecasts, while Ezzat et al.~\citep{ezzat2019spatio} proposed calibrated regime-switching methods that correct systematic biases within each regime. These works establish that conditioning the forecast on regime-specific parameters is a principled and effective strategy, but they rely on \emph{predefined} regime definitions and separately fitted models.

\paragraph{Clustered deep learning approaches}
More recently, several studies have combined clustering with deep neural networks to extend the regime-switching paradigm. Zhang et al.~\citep{zhang2020short} grouped NWP-derived weather features via clustering and trained a dedicated Seq2Seq model for each cluster. Wang et al.~\citep{wang2018deep} used a deep belief network followed by $k$-means clustering to partition the input space, and Jiang et al.~\citep{jiang2024novel} integrated feature selection with neural network clustering and BiGRU for per-cluster multi-step prediction. These approaches demonstrate the benefit of regime-dependent modeling within neural architectures, yet they share a common limitation: the clustering is performed as a \emph{preprocessing step} and remains fixed during downstream training, preventing the regime assignments from adapting to the forecasting objective. Moreover, training a separate model for each cluster increases the total number of parameters and prevents cross-regime representation sharing.

\paragraph{ECTO's position}
ECTO departs from both classical and cluster-then-predict approaches by learning regime routing \emph{inside} a unified model. Rather than pre-clustering and training separate predictors, the ECRR module dynamically generates soft regime gates conditioned on the exogenous context selected by PGVS, and applies regime-specific gain and bias adjustments to the base forecast. This end-to-end formulation allows the regime structure to emerge as a by-product of the training objective and to adapt jointly with the variable selection and temporal encoding modules.

\section{Methodology}
\label{sec:method}

\paragraph{Notation}
Throughout this section, scalars are denoted by lowercase letters ($x$, $\alpha$), vectors by bold lowercase ($\mathbf{x}$, $\mathbf{h}$), and matrices by bold uppercase ($\mathbf{X}$, $\mathbf{T}$). The operator $[\,\cdot \,\|\, \cdot\,]$ denotes concatenation along the last dimension. The $\ell_p$ norm is written as $\|\cdot\|_p$. Subscripts index time steps ($x_t$), variables ($\mathbf{s}_d$), groups ($\mathcal{G}_g$), regime experts ($k$), or horizon steps ($t$). Superscripts distinguish semantic roles: ``tar'' (target), ``exo'' (exogenous), ``base'' (base forecast), and ``regime'' (regime-refined). Key dimensions are: input length $T$, prediction horizon $H$, number of exogenous variables $D$, number of physical groups $G$, token embedding dimension $d$, exogenous embedding dimension $d_e$, target state dimension $d_s$, horizon embedding dimension $d_h$, and number of regime experts $K$.

\subsection{Problem Formulation}
\label{sec:problem}

We consider the task of ultra-short-term wind power forecasting, where the goal is to predict future wind power generation over the next $H$ time steps given historical observations of both the target power variable and a set of exogenous meteorological variables.

\paragraph{Input}
Let $\mathbf{x}^{\mathrm{tar}} = \{x^{\mathrm{tar}}_1, x^{\mathrm{tar}}_2, \ldots, x^{\mathrm{tar}}_T\} \in \mathbb{R}^{T}$ denote the historical target sequence (wind power) over a lookback window of $T$ time steps, and let $\mathbf{X}^{\mathrm{exo}} = \{\mathbf{x}^{\mathrm{exo}}_1, \mathbf{x}^{\mathrm{exo}}_2, \ldots, \mathbf{x}^{\mathrm{exo}}_T\} \in \mathbb{R}^{T \times D}$ denote the corresponding $D$ exogenous variables over the same window. Each column of $\mathbf{X}^{\mathrm{exo}}$ corresponds to a meteorological variable such as wind speed at different heights, wind direction, temperature, pressure, or humidity. The full input is the multivariate time series $\mathbf{X} = [\mathbf{X}^{\mathrm{exo}} \, \| \, \mathbf{x}^{\mathrm{tar}}] \in \mathbb{R}^{T \times (D+1)}$.

\paragraph{Output}
The forecasting objective is to produce $\hat{\mathbf{y}} = \{\hat{y}_1, \hat{y}_2, \ldots, \hat{y}_H\} \in \mathbb{R}^{H}$, an estimate of future wind power over the prediction horizon $H$.

\paragraph{Asymmetric roles}
A distinguishing feature of this problem is the \emph{asymmetric role} of the target and exogenous variables. The target variable $\mathbf{x}^{\mathrm{tar}}$ is the quantity to be predicted; its temporal patterns carry the primary predictive signal. The exogenous variables $\mathbf{X}^{\mathrm{exo}}$ serve as \emph{conditional context}---they modulate and refine the target-based forecast but do not themselves require prediction. This asymmetry motivates an architecture that treats the two modalities differently rather than processing all channels uniformly.

\paragraph{Physical grouping prior}
The $D$ exogenous variables are partitioned into $G$ physically motivated groups $\{\mathcal{G}_1, \ldots, \mathcal{G}_G\}$ (with $\mathcal{G}_g \subset \{1, \ldots, D\}$ and $\bigcup_g \mathcal{G}_g = \{1, \ldots, D\}$). Unlike generic multivariate time series, meteorological variables in wind power forecasting exhibit clear physical dependencies: for example, wind speeds at different heights are coupled by the power law, while temperature and pressure jointly determine air density. This domain knowledge motivates a structured variable selection that exploits such physical groupings rather than treating all inputs uniformly. The specific grouping strategy is detailed in Section~\ref{sec:pgvs}.

\paragraph{Input normalization}
Because wind power time series exhibit strong distributional shifts across seasons and operating conditions, each input window is normalized by subtracting its temporal mean and dividing by its standard deviation before entering the model. At inference time, the prediction is denormalized using the statistics of the input window, under the assumption that the forecast horizon shares the same distributional shift~\citep{kim2022revin}.

\paragraph{Objective}
The forecasting model $f_\theta$ is trained to minimize the mean squared error between the predicted and true future power values:
\begin{equation}
  \min_{\theta} \; \mathcal{L}(\mathbf{y}, \hat{\mathbf{y}}) = \min_{\theta} \; \frac{1}{H} \sum_{h=1}^{H} (y_h - \hat{y}_h)^2,
  \label{eq:objective}
\end{equation}
where $\mathbf{y} \in \mathbb{R}^{H}$ denotes the ground-truth future power and $\hat{\mathbf{y}} = f_\theta(\mathbf{x}^{\mathrm{tar}}, \mathbf{X}^{\mathrm{exo}})$.

\subsection{Architecture Overview}
\label{sec:overview}

\begin{figure}[htbp]
\centering
\includegraphics[width=\textwidth,height=0.85\textheight,keepaspectratio,clip,trim=0 0 0 0]{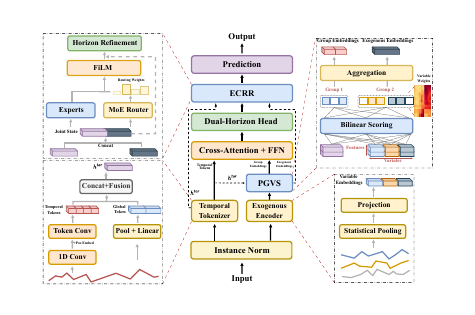}
\caption{Overall architecture of ECTO. The target power sequence is encoded by the Temporal Tokenizer into Temporal Tokens ($\mathbf{T}$) and a global state vector ($\mathbf{h}^{\mathrm{tar}}$). The Exogenous Encoder compresses each exogenous variable via Statistical Pooling and Projection into Variable Embeddings ($\mathbf{S}$). PGVS performs hierarchical selection over these embeddings via Bilinear Scoring, producing Group Embeddings ($\mathbf{C}$, for cross-attention fusion) and Exogenous Embeddings ($\mathbf{z}^{\mathrm{exo}}$, for downstream conditioning). A Cross-Attention layer fuses Group Embeddings into the target tokens, and a Dual-Horizon Head produces the base forecast. ECRR then consumes both $\mathbf{h}^{\mathrm{tar}}$ and $\mathbf{z}^{\mathrm{exo}}$ to produce the final prediction. Dashed lines indicate conditioning signals; solid lines indicate primary data flow. The inset heatmap shows learned PGVS variable weights across power levels (WF1).}
\label{fig:architecture}
\end{figure}

Fig.~\ref{fig:architecture} illustrates the overall architecture of ECTO. Given the input pair $(\mathbf{x}^{\mathrm{tar}}, \mathbf{X}^{\mathrm{exo}})$, ECTO decomposes exogenous variable modeling into two complementary modules that address distinct aspects of the forecasting problem: \emph{which} exogenous variables matter for the current prediction, and \emph{how} they should modulate the forecast under different conditions.

\textbf{Target encoding.}
The target sequence $\mathbf{x}^{\mathrm{tar}} \in \mathbb{R}^{T}$ is encoded by a \emph{Temporal Tokenizer} (Section~\ref{sec:tokenizer}) into a sequence of local tokens $\mathbf{T} \in \mathbb{R}^{N \times d}$ and a global state vector $\mathbf{h}^{\mathrm{tar}} \in \mathbb{R}^{d_s}$. Here $\mathbf{T}$ captures local temporal patterns (ramps, peaks, plateaus) while $\mathbf{h}^{\mathrm{tar}}$ summarizes the overall target condition.

\textbf{PGVS: Which exogenous variables matter?}
The exogenous variables $\mathbf{X}^{\mathrm{exo}} \in \mathbb{R}^{T \times D}$ are first encoded into variable-level representations $\mathbf{S} \in \mathbb{R}^{D \times d_e}$ via statistical summarization. PGVS then performs a \emph{hierarchical selection}---screening physical groups via sparse gating, then selecting informative variables within each group---producing a compact group-level context $\mathbf{C} \in \mathbb{R}^{G \times d_e}$. A cross-attention layer fuses this context into the target tokens: $\mathbf{T}^{\mathrm{fused}} = \mathrm{CrossAttn}(\mathbf{T}, \mathbf{C})$, yielding a target representation enriched with selected exogenous information. A dual-horizon prediction head maps $\mathbf{T}^{\mathrm{fused}}$ to a base forecast $\hat{\mathbf{y}}^{\mathrm{base}} \in \mathbb{R}^{H}$ (Sections~\ref{sec:exo-encode}--\ref{sec:cross-attn}).

\textbf{ECRR: When and how should exogenous variables affect the forecast?}
Operating on the base forecast, ECRR conditions a \emph{regime-aware refinement} on the exogenous context. A regime router maps the concatenated state $[\mathbf{h}^{\mathrm{tar}}; \mathbf{z}^{\mathrm{exo}}]$---where $\mathbf{z}^{\mathrm{exo}} = \sum_g w_g \mathbf{c}_g$ aggregates the group contexts using the learned group-importance weights $w_g$ from PGVS---to a distribution over $K$ regimes (Section~\ref{sec:regime-router}). Each regime expert produces gain and bias adjustments, which are softly mixed and applied to the base forecast via gated multiplicative-additive modulation (Section~\ref{sec:gain-bias}). A horizon-specific refinement further adapts the correction for each prediction step, yielding the final output $\hat{\mathbf{y}} \in \mathbb{R}^{H}$ (Section~\ref{sec:horizon}).

In summary, PGVS addresses the \emph{selection} question---identifying which exogenous variables carry predictive signal for each input sample---while ECRR addresses the \emph{modulation} question---determining how the selected variables should conditionally adjust the forecast under different operating regimes.

\subsection{Temporal Tokenizer}
\label{sec:tokenizer}

The Temporal Tokenizer converts the univariate target sequence $\mathbf{x}^{\mathrm{tar}} \in \mathbb{R}^{T}$ into two complementary representations: a sequence of local temporal tokens $\mathbf{T} \in \mathbb{R}^{N \times d}$ that preserve fine-grained temporal structure, and a global state vector $\mathbf{h}^{\mathrm{tar}} \in \mathbb{R}^{d_s}$ that summarizes the overall target condition. Both representations serve as the target-side anchors for subsequent exogenous variable selection and regime-aware refinement.

\paragraph{Hierarchical convolutional tokenization}
Recent patch-based forecasting methods~\citep{nie2023patchtst} segment the input into disjoint patches and project each patch linearly. While effective for capturing local semantics, a single linear projection cannot detect structured local patterns such as ramps, peaks, or sustained plateaus that are characteristic of wind power signals. Instead, we employ a two-stage hierarchical convolution:
\begin{align}
  \mathbf{H} &= \mathrm{Conv1d}_{\mathrm{local}}(\mathbf{x}^{\mathrm{tar}}; \, c \text{ filters}, k_1, \text{same padding}) \in \mathbb{R}^{c \times T}, \label{eq:local-conv} \\
  \mathbf{T} &= \mathrm{Conv1d}_{\mathrm{token}}(\mathbf{H}; \, c \text{ input}, d \text{ output}, k_2, s)^{\top} \in \mathbb{R}^{N \times d}. \label{eq:token-conv}
\end{align}
The first convolution applies $c$ local feature detectors (kernel size $k_1$) with same padding to preserve the temporal resolution $T$, where each filter learns to detect a distinct temporal motif. The second convolution (kernel size $k_2$, stride $s < k_2$) simultaneously segments the feature maps into overlapping patches and projects them into $d$-dimensional token embeddings. The overlap ($s < k_2$) ensures that adjacent tokens share receptive fields, preserving temporal continuity at patch boundaries. The resulting token count is $N = \lfloor (T + 2p - k_2) / s \rfloor + 1$ with symmetric padding $p = k_2 / 2$ ($k_2$ is even by design).

\paragraph{Global state vector}
The target state $\mathbf{h}^{\mathrm{tar}}$ is constructed from two learned representations and a lightweight statistical summary. First, we pool the local feature map $\mathbf{H}$ by concatenating its temporal mean, max, and last step, and project the result to dimension $d$:
\begin{equation}
  \mathbf{h}^{\mathrm{pool}} = \mathrm{Linear}\bigl([\,\mathrm{mean}(\mathbf{H}) \,\|\, \mathrm{max}(\mathbf{H}) \,\|\, \mathbf{H}_{:,-1}\,]\bigr) \in \mathbb{R}^{d}.
\end{equation}
Second, we compute the mean-pooled token representation $\bar{\mathbf{T}} = \frac{1}{N}\sum_n \mathbf{T}_n \in \mathbb{R}^{d}$. These two learned vectors are concatenated with a small set of descriptive statistics of $\mathbf{x}^{\mathrm{tar}}$ (e.g., mean, standard deviation, range, short-window trend) and passed through a two-layer MLP to produce the final target state:
\begin{equation}
  \mathbf{h}^{\mathrm{tar}} = \mathrm{MLP}\bigl([\,\mathbf{h}^{\mathrm{pool}} \,\|\, \bar{\mathbf{T}} \,\|\, \mathbf{s}^{\mathrm{tar}}\,]\bigr) \in \mathbb{R}^{d_s}.
\end{equation}
This state vector serves as the conditioning signal for both PGVS group scoring and ECRR regime routing.

\paragraph{Positional encoding}
Learnable positional embeddings $\mathbf{P} \in \mathbb{R}^{N \times d}$ are added to the token sequence, yielding $\mathbf{T} = \mathbf{T} + \mathbf{P}$, so that the downstream cross-attention can distinguish tokens from different temporal positions.

\subsection{PGVS: Physical Group-aware Exogenous Variable Selection}
\label{sec:pgvs}

PGVS addresses the question of \emph{which} exogenous variables carry predictive signal for the current sample. It operates in four stages: encoding each exogenous variable into a compact representation, scoring physical groups via sparse gating, selecting informative variables within each group, and fusing the selected exogenous context into the target tokens via cross-attention.

\subsubsection{Exogenous Encoding}
\label{sec:exo-encode}

The raw exogenous time series $\mathbf{X}^{\mathrm{exo}} \in \mathbb{R}^{T \times D}$ contains high-dimensional temporal detail that is redundant for variable-level decision-making. We compress each exogenous variable into a fixed-length descriptor by computing statistics over two temporal scales---a short window of $T_s$ steps (capturing recent dynamics) and a long window of $T_l$ steps (capturing broader trends):
\begin{equation}
  \mathbf{f}_d = \bigl[\,\mu_d^{(s)},\; \sigma_d^{(s)},\; x_{d,T}^{(s)},\; \mu_d^{(l)},\; \sigma_d^{(l)},\; x_{d,T}^{(l)}\,\bigr] \in \mathbb{R}^{6}, \quad d = 1, \ldots, D,
  \label{eq:exo-stats}
\end{equation}
where $\mu_d^{(s)}, \sigma_d^{(s)}$ denote the mean and standard deviation of variable $d$ over the short window, and $x_{d,T}^{(s)}$ its last value (similarly for the long window). These $D$ descriptor vectors are stacked and passed through a shared two-layer MLP:
\begin{equation}
  \mathbf{S} = \bigl[\,\mathbf{s}_1;\, \ldots;\, \mathbf{s}_D\,\bigr], \quad \mathbf{s}_d = \mathrm{MLP}_{\mathrm{exo}}(\mathbf{f}_d) \in \mathbb{R}^{d_e}, \quad \mathbf{S} \in \mathbb{R}^{D \times d_e}.
  \label{eq:s-exo}
\end{equation}
This encoding reduces the temporal dimension from $T$ to $d_e$ per variable while preserving the multi-scale information needed for subsequent selection.

\subsubsection{Physical Grouping Prior}
\label{sec:group-prior}

The $D$ exogenous variables are partitioned into $G$ physically motivated groups $\{\mathcal{G}_1, \ldots, \mathcal{G}_G\}$, as defined in Section~\ref{sec:problem}. For the wind power datasets in this study, we use $G = 2$: a \emph{wind speed group} containing multi-height wind speed measurements, and an \emph{atmospheric group} containing temperature, pressure, humidity, and wind direction variables. This grouping reflects the physical hierarchy of the atmospheric boundary layer---wind speed is the primary driver of power conversion, while other atmospheric conditions act as secondary modulators. For each group, we compute a summary vector by averaging its member representations:
\begin{equation}
  \bar{\mathbf{s}}_g = \frac{1}{|\mathcal{G}_g|} \sum_{d \in \mathcal{G}_g} \mathbf{s}_d \in \mathbb{R}^{d_e}, \quad g = 1, \ldots, G.
  \label{eq:group-summary}
\end{equation}

\subsubsection{Group Screening}
\label{sec:group-screen}

Not all physical groups are equally informative for every prediction. PGVS assigns an importance weight to each group through a target-conditioned scoring function followed by sparse gating. The scoring function uses a bilinear interaction between the target state and each group summary:
\begin{equation}
  \pi_g = \mathbf{v}^{\top} \tanh\!\bigl(\mathbf{W}_h \mathbf{h}^{\mathrm{tar}} + \mathbf{W}_s \bar{\mathbf{s}}_g + \mathbf{W}_{hs} (\mathbf{W}_{\mathrm{proj}} \mathbf{h}^{\mathrm{tar}} \odot \bar{\mathbf{s}}_g)\bigr),
  \label{eq:group-score}
\end{equation}
where $\mathbf{W}_h, \mathbf{W}_s \in \mathbb{R}^{r \times \cdot}$ are projection matrices, $\mathbf{W}_{hs}$ captures the target--exogenous interaction, and $\mathbf{v} \in \mathbb{R}^{r}$ produces a scalar score. The raw scores are scaled by a temperature $\tau_g$ and passed through a \emph{sparsemax}~\citep{martins2016softmax} activation:
\begin{equation}
  \mathbf{w}^G = \mathrm{sparsemax}\!\left(\frac{\boldsymbol{\pi}}{\tau_g}\right) \in \mathbb{R}^{G},
  \label{eq:group-weight}
\end{equation}
where $\boldsymbol{\pi} = [\pi_1, \ldots, \pi_G]$ and $\tau_g$ is a temperature hyperparameter. Unlike softmax, sparsemax can produce exact zeros, enabling the model to entirely suppress irrelevant groups. This is particularly desirable in our setting: under certain weather conditions, one physical group may carry negligible predictive information, and sparse gating avoids diluting the signal with near-uniform weights. Moreover, by resolving relevance at the group level first, the subsequent variable-level selection operates within a reduced and physically coherent subspace rather than searching over all variables independently.

\subsubsection{Variable Selection and Exogenous Context}
\label{sec:var-select}

Within each surviving group, PGVS further selects the most informative variables. The same bilinear scoring architecture (Eq.~\ref{eq:group-score}) is applied at the variable level, with each individual variable representation $\mathbf{s}_d$ replacing the group summary $\bar{\mathbf{s}}_g$ in the interaction terms and a separate set of parameters, producing logits $\pi_d^{\mathrm{var}}$ for each variable $d$. Within group $\mathcal{G}_g$, these logits are normalized via softmax:
\begin{equation}
  \alpha_d = \frac{\exp(\pi_d^{\mathrm{var}} / \tau_v)}{\sum_{d' \in \mathcal{G}_g} \exp(\pi_{d'}^{\mathrm{var}} / \tau_v)}, \quad d \in \mathcal{G}_g,
  \label{eq:var-weight}
\end{equation}
where $\tau_v$ is a separate temperature. Optionally, a top-$k$ mask retains only the $k$ highest-scoring variables per group and renormalizes. The group-level and variable-level weights are combined into a \emph{hierarchical weight}:
\begin{equation}
  w_d^{\mathrm{hier}} = w_g^G \cdot \alpha_d, \quad d \in \mathcal{G}_g.
  \label{eq:hier-weight}
\end{equation}
This multiplicative decomposition ensures that a variable can receive a high weight only if both its group is deemed relevant and the variable itself ranks highly within that group, enforcing a physically structured selection hierarchy rather than an unconstrained search over all variables.
These weights yield two outputs. First, a \emph{group-level context}---a compact representation of the selected exogenous information at the group level:
\begin{equation}
  \mathbf{c}_g = w_g^G \cdot \sum_{d \in \mathcal{G}_g} \alpha_d \, \mathbf{s}_d \in \mathbb{R}^{d_e}, \quad \mathbf{C} = [\mathbf{c}_1; \ldots; \mathbf{c}_G] \in \mathbb{R}^{G \times d_e}.
  \label{eq:group-context}
\end{equation}
Second, a \emph{scalar exogenous summary} obtained by aggregating the group summaries with the group weights:
\begin{equation}
  \mathbf{z}^{\mathrm{exo}} = \sum_{g=1}^{G} w_g^G \, \bar{\mathbf{s}}_g \in \mathbb{R}^{d_e},
  \label{eq:z-exo}
\end{equation}
which will serve as the exogenous input to the ECRR module (Section~\ref{sec:ecrr}). Together, $\mathbf{C}$ and $\mathbf{z}^{\mathrm{exo}}$ form a two-level information flow: $\mathbf{C}$ carries variable-level detail for cross-attention fusion with target tokens, while $\mathbf{z}^{\mathrm{exo}}$ provides a compact, group-level summary for downstream regime conditioning.

\subsubsection{Target-Exogenous Context Fusion}
\label{sec:cross-attn}

The group-level context $\mathbf{C}$ is projected to the target embedding dimension and fused into the target tokens via cross-attention~\citep{vaswani2017attention}:
\begin{align}
  \mathbf{C}' &= \mathrm{Linear}(\mathbf{C}) \in \mathbb{R}^{G \times d}, \label{eq:group-token} \\
  \mathbf{A} &= \mathrm{CrossAttn}\!\bigl(\mathbf{Q}=\mathrm{LN}(\mathbf{T}),\; \mathbf{K}=\mathbf{C}',\; \mathbf{V}=\mathbf{C}'\bigr), \label{eq:cross-attn} \\
  \mathbf{T}^{\mathrm{fused}} &= \mathbf{T} + \sigma(\eta) \cdot \mathrm{Proj}\!\bigl(\mathbf{A}\bigr) + \mathrm{FFN}\!\bigl(\mathbf{T} + \sigma(\eta) \cdot \mathrm{Proj}\!\bigl(\mathbf{A}\bigr)\bigr), \label{eq:token-fused}
\end{align}
where $\eta$ is a learnable scalar initialized to a small negative value so that $\sigma(\eta) \approx 0$ at the start of training, $\mathrm{Proj}$ is a two-layer projection with intermediate activation, and $\mathrm{FFN}$ is a standard feed-forward block. The gated residual $\sigma(\eta)$ ensures that exogenous information is injected conservatively, preventing it from overwhelming the target signal. The resulting $\mathbf{T}^{\mathrm{fused}} \in \mathbb{R}^{N \times d}$ encodes target temporal patterns enriched with selected exogenous context.

A dual-horizon prediction head then flattens $\mathbf{T}^{\mathrm{fused}}$ and produces the base forecast:
\begin{equation}
  \hat{\mathbf{y}}^{\mathrm{base}} = \bigl[\,\mathrm{Linear}_{\mathrm{short}}(\mathrm{flatten}(\mathbf{T}^{\mathrm{fused}})) \;\|\; \mathrm{Linear}_{\mathrm{long}}(\mathrm{flatten}(\mathbf{T}^{\mathrm{fused}}))\,\bigr] \in \mathbb{R}^{H},
  \label{eq:base-pred}
\end{equation}
where the short head predicts the first $H/2$ steps and the long head predicts the remaining $H/2$ steps, allowing the model to specialize its representation for different forecast horizons. Each head consists of a dropout layer followed by a single linear projection, mapping directly from the flattened token representation to the corresponding horizon steps.

\subsection{ECRR: Exogenous-Conditioned Regime Refinement}
\label{sec:ecrr}

While PGVS determines which exogenous variables to incorporate, ECRR addresses how they should modulate the forecast. Wind power generation exhibits distinct predictive regimes---stable periods, ramp events, and curtailed operation---where the mapping from input features to power output follows different patterns~\citep{browell2018improved,ezzat2019spatio}. Rather than pre-clustering samples and training separate predictors, ECRR adopts a mixture-of-experts (MoE) paradigm~\citep{jacobs1991moe} that learns exogenous-conditioned regime routing inside a unified refinement module. A lightweight router maps the exogenous context to a soft assignment over $K$ regime experts, each producing a gain--bias adjustment; the expert outputs are then softly blended and applied to the base forecast via gated multiplicative-additive modulation, followed by a horizon-specific refinement.

\subsubsection{Exogenous-Conditioned Regime Router}
\label{sec:regime-router}

The regime router takes as input the concatenation of the target state and the exogenous summary from PGVS:
\begin{equation}
  \mathbf{q} = [\,\mathbf{h}^{\mathrm{tar}} \,\|\, \mathbf{z}^{\mathrm{exo}}\,] \in \mathbb{R}^{d_s + d_e},
  \label{eq:calib-in}
\end{equation}
where $\mathbf{z}^{\mathrm{exo}}$ (Eq.~\ref{eq:z-exo}) carries the group-level exogenous context. A linear router produces a soft assignment over $K$ regimes:
\begin{equation}
  \mathbf{r} = \mathrm{softmax}\!\bigl(\mathbf{W}_r \mathbf{q} + \mathbf{b}_r\bigr) \in \mathbb{R}^{K},
  \label{eq:regime-weight}
\end{equation}
where $\mathbf{r} = [r_1, \ldots, r_K]$ with $\sum_k r_k = 1$. Crucially, the routing depends on the exogenous context $\mathbf{z}^{\mathrm{exo}}$: two samples with similar target histories but different atmospheric conditions may be routed to different regimes, allowing the model to adapt its refinement strategy to the prevailing weather regime.

\subsubsection{Expert Gain/Bias Modulation}
\label{sec:gain-bias}

For each regime expert $k$, a dedicated linear head produces a per-horizon gain and bias from the conditioning vector $\mathbf{q}$:
\begin{align}
  \mathbf{g}_k &= \mathbf{W}^g_k \mathbf{q} + \mathbf{b}^g_k \in \mathbb{R}^{H}, \label{eq:regime-gain-raw} \\
  \mathbf{b}_k &= \mathbf{W}^b_k \mathbf{q} + \mathbf{b}^b_k \in \mathbb{R}^{H}. \label{eq:regime-bias-raw}
\end{align}
The regime-specific outputs are mixed according to the routing weights:
\begin{equation}
  \mathbf{g}^{\mathrm{mix}} = \sum_{k=1}^{K} r_k \, \mathbf{g}_k, \quad \mathbf{b}^{\mathrm{mix}} = \sum_{k=1}^{K} r_k \, \mathbf{b}_k.
  \label{eq:regime-mix}
\end{equation}
To prevent aggressive modification of the base prediction early in training, each output is passed through a sigmoid-gated FiLM~\citep{perez2018film} mechanism:
\begin{equation}
  \hat{\mathbf{y}}^{\mathrm{regime}} = \bigl(1 + \sigma(\boldsymbol{\gamma}_g) \odot \tanh(\mathbf{g}^{\mathrm{mix}})\bigr) \odot \hat{\mathbf{y}}^{\mathrm{base}} + \sigma(\boldsymbol{\gamma}_b) \odot \mathbf{b}^{\mathrm{mix}},
  \label{eq:regime-pred}
\end{equation}
where $\boldsymbol{\gamma}_g, \boldsymbol{\gamma}_b \in \mathbb{R}^{H}$ are learnable gate parameters initialized to small negative values so that $\sigma(\boldsymbol{\gamma}) \approx 0$ at the start of training. The $\tanh$ bounds the raw gain logits to $[-1, 1]$, preventing extreme multiplicative factors when the gate opens. This conservative initialization ensures that ECRR begins as a near-identity mapping and gradually learns to apply regime-specific adjustments as training progresses. The gain term $(1 + \cdot)$ performs multiplicative scaling (amplifying or attenuating the base forecast), while the bias term performs additive shifting---together capturing the observation that prediction errors under different weather regimes exhibit both scale and offset biases.

\subsubsection{Horizon-Specific Refinement}
\label{sec:horizon}

Prediction errors typically vary across forecast horizons---near-term steps are dominated by persistence while later steps accumulate atmospheric uncertainty. ECRR applies a second refinement stage that operates at the per-horizon level. A context vector is first extracted from $\mathbf{q}$ and combined with a learnable horizon embedding $\mathbf{E}_h \in \mathbb{R}^{H \times d_h}$ through a multiplicative interaction:
\begin{align}
  \mathbf{h}^{\mathrm{ctx}} &= \mathrm{MLP}_{\mathrm{ctx}}(\mathbf{q}) \in \mathbb{R}^{d_h}, \label{eq:hctx} \\
  \mathbf{h}^{\mathrm{feat}}_t &= \mathbf{h}^{\mathrm{ctx}} + \mathbf{E}_{h,t} + \mathbf{h}^{\mathrm{ctx}} \odot \mathbf{E}_{h,t}, \quad t = 1, \ldots, H, \label{eq:hfeat}
\end{align}
where $\mathbf{E}_{h,t}$ is the embedding for horizon step $t$. The multiplicative term $\mathbf{h}^{\mathrm{ctx}} \odot \mathbf{E}_{h,t}$ allows the refinement magnitude to depend on the interaction between the current exogenous conditions and the specific horizon position. A shared lightweight MLP is applied independently to each $\mathbf{h}^{\mathrm{feat}}_t$ to produce a scalar correction:
\begin{equation}
  \delta_t = \mathrm{MLP}_{\delta}(\mathbf{h}^{\mathrm{feat}}_t) \in \mathbb{R}, \quad t = 1, \ldots, H,
  \label{eq:delta-h}
\end{equation}
and the per-step corrections are stacked into $\boldsymbol{\delta}_h = [\delta_1, \ldots, \delta_H] \in \mathbb{R}^{H}$,
which is gated and added to the regime-refined prediction:
\begin{equation}
  \hat{\mathbf{y}} = \hat{\mathbf{y}}^{\mathrm{regime}} + \sigma(\boldsymbol{\gamma}_h) \odot \boldsymbol{\delta}_h,
  \label{eq:final-pred}
\end{equation}
where $\boldsymbol{\gamma}_h$ is a learnable gate similarly initialized near zero. The final output $\hat{\mathbf{y}} \in \mathbb{R}^{H}$ is the refined prediction that incorporates both regime-level and horizon-level adjustments, both conditioned on the exogenous context from PGVS.

\subsection{Training Objective}
\label{sec:training}

The model is trained by minimizing a composite loss that combines mean squared error (MSE) with a weighted mean absolute error (MAE) term:
\begin{equation}
  \mathcal{L} = \frac{1}{B \cdot H} \sum_{i=1}^{B} \bigl\|\hat{\mathbf{y}}_i - \mathbf{y}_i\bigr\|_2^2 + \lambda \cdot \frac{1}{B \cdot H} \sum_{i=1}^{B} \bigl\|\hat{\mathbf{y}}_i - \mathbf{y}_i\bigr\|_1,
  \label{eq:loss}
\end{equation}
where $B$ is the batch size, $\mathbf{y}_i \in \mathbb{R}^{H}$ is the ground truth, and $\lambda$ is a weighting coefficient. The MAE term provides a constant-gradient component that remains effective for small residuals, while the MSE term accelerates convergence on large errors.

\section{Experiments}
\label{sec:exp}

\subsection{Experimental Setup}
\label{sec:setup}

\subsubsection{Datasets}
\label{sec:datasets}

We evaluate ECTO on three wind farm datasets that differ in rated capacity, climate zone, and data distribution. Table~\ref{tab:datasets} summarizes the key characteristics.

\begin{table}[htbp]
\centering
\caption{Summary of the three wind farm datasets. The $D$ exogenous variables consist of multi-height wind speed, wind direction, temperature, pressure, and humidity.}
\label{tab:datasets}
\resizebox{\textwidth}{!}{
\begin{tabular}{lccc}
\toprule
& \textbf{WF1} & \textbf{WF4} & \textbf{Xinjiang} \\
\midrule
Rated capacity (MW) & 99 & 66 & 200 \\
Region & Eastern China & Eastern China & Northwest China \\
Time span & 2019--2020 & 2019--2020 & 2019 \\
Resolution & 15 min & 15 min & 15 min \\
Total samples & 70,176 & 70,176 & 35,040 \\
Exogenous vars.\ ($D$) & 11 & 11 & 13 \\
Wind speed heights & 4 (10/30/50/hub) & 4 (10/30/50/hub) & 5 (10/30/50/70/hub) \\
Atm.\ other & temp., pres., hum. & temp., pres., hum. & temp., pres., hum. \\
Zero-power ratio & 0.29\% & 19.69\% & $<$0.01\% \\
Capacity factor & 23.7\% & 26.3\% & 35.1\% \\
Data source & \cite{chen2022stategrid} & \cite{chen2022stategrid} & Private \\
\bottomrule
\end{tabular}
}
\end{table}

\textbf{WF1} and \textbf{WF4} are drawn from the open dataset published by Chen and Xu~\cite{chen2022stategrid}, which covers six wind farms across different climate zones in China with two years of SCADA measurements at 15-minute intervals. WF1 has a rated capacity of 99\,MW and an exceptionally low zero-power ratio of 0.29\%, indicating near-continuous operation and making it well suited for evaluating prediction accuracy under active generation. WF4 has a rated capacity of 66\,MW and a significantly higher zero-power ratio of 19.69\%, reflecting more frequent shutdowns or curtailment events. Notably, despite its higher zero-power ratio, WF4 exhibits a slightly higher capacity factor than WF1 (26.3\% vs.\ 23.7\%), suggesting that generation during active periods is more efficient and that zero-power events are concentrated in low-wind intervals. This decoupling of availability and efficiency makes WF4 a valuable stress test for model robustness under substantially different power distributions.

\textbf{Xinjiang} is collected from a 200\,MW wind farm in the arid northwest region of China and is not publicly available. It provides one year of measurements at 15-minute resolution. Compared with WF1 and WF4, this site features five layers of wind speed and direction measurements (adding a 70\,m layer), a higher capacity factor (35.1\%), and a distinct continental climate with large diurnal temperature swings exceeding 25\,$^{\circ}$C, which challenge the model's ability to condition on rapidly evolving atmospheric states. Its inclusion tests whether the proposed physical grouping prior and exogenous conditioning mechanism transfer across different geographic and meteorological conditions.

For all three datasets, each sample comprises the target variable (wind power) together with exogenous variables at multiple height levels. The data are chronologically split into training (70\%), validation (10\%), and test (20\%) sets without shuffling, which preserves temporal order and prevents information leakage from future observations into the training set, reflecting the real-world forecasting setting where models are trained on historical data and deployed on future data. Missing values are imputed via linear interpolation with a maximum gap of 32 consecutive steps (8 hours); any remaining gaps are filled by forward- and backward-filling with a limit of 8 steps. After preprocessing, all datasets contain no missing values.

\subsubsection{Baselines}
\label{sec:baselines}

ECTO is compared against nine representative time-series forecasting models spanning different architectural paradigms, plus a persistence baseline:

\begin{itemize}
\item \textbf{Persistence}: A naive baseline that uses the last observed value as the forecast for all horizons, serving as a lower-bound reference for forecast difficulty.
\item \textbf{Informer}~\cite{zhou2021informer}: A Transformer variant with ProbSparse self-attention designed for long-sequence forecasting.
\item \textbf{DLinear}~\cite{zeng2023transformers}: A lightweight model that decomposes the input into trend and seasonal components and applies independent linear layers, serving as a strong baseline that questions the necessity of attention mechanisms.
\item \textbf{PatchTST}~\cite{nie2023patchtst}: A Transformer model that segments the input into patches to reduce sequence length and enlarge the receptive field, achieving state-of-the-art results on multivariate benchmarks.
\item \textbf{TimesNet}~\cite{wu2023timesnet}: A model that transforms 1D time series into 2D tensors and applies 2D convolutions to capture intra- and inter-period variations.
\item \textbf{iTransformer}~\cite{liu2024itransformer}: An inverted Transformer that treats each variate as an independent token and applies attention across the variate dimension rather than the temporal dimension.
\item \textbf{TimeMixer}~\cite{wang2024timemixer}: A fully MLP-based multi-scale mixing model that leverages past series decomposition and future multipredictor mixing across multiple resolutions.
\item \textbf{TimeXer}~\cite{wang2024timexer}: A Transformer model that incorporates exogenous variables by injecting endogenous patch tokens with external information through cross-attention.
\item \textbf{XLinear}~\cite{chen2026xlinear}: An MLP-based model with variable-wise gating that explicitly models the relationship between exogenous variables and the target, representing a strong recent baseline for exogenous-aware forecasting.
\item \textbf{CrossLinear}~\cite{zhou2025crosslinear}: A lightweight model with a plug-and-play cross-correlation embedding module that captures direct and time-invariant dependencies between exogenous and endogenous variables via 1D convolution.
\end{itemize}

This selection covers the spectrum from attention-based models (Informer, PatchTST, iTransformer) to lightweight architectures (DLinear, XLinear, CrossLinear), and includes three models that explicitly handle exogenous variables (TimeXer, XLinear, CrossLinear), enabling a direct assessment of the proposed exogenous conditioning mechanism.

\subsubsection{Evaluation Metrics}
\label{sec:metrics}

We adopt three standard metrics for time-series forecasting. Let $\hat{y}_t$ and $y_t$ denote the predicted and observed values, respectively, and $\bar{y}$ the sample mean of the ground truth. MAE and NSE are reported as complementary metrics to assess bias and overall predictive skill, respectively, while MSE serves as the primary metric for model selection and comparison.

\paragraph{Mean Squared Error (MSE)}
\begin{equation}
\mathrm{MSE} = \frac{1}{H}\sum_{t=1}^{H}(\hat{y}_t - y_t)^2
\end{equation}
MSE penalizes large errors quadratically.

\paragraph{Mean Absolute Error (MAE)}
\begin{equation}
\mathrm{MAE} = \frac{1}{H}\sum_{t=1}^{H}|\hat{y}_t - y_t|
\end{equation}
MAE provides a linear penalty on residuals and is less sensitive to outliers than MSE.

\paragraph{Nash--Sutcliffe Efficiency (NSE)}
\begin{equation}
\mathrm{NSE} = 1 - \frac{\sum_{t=1}^{H}(\hat{y}_t - y_t)^2}{\sum_{t=1}^{H}(y_t - \bar{y})^2}
\end{equation}
NSE measures the fraction of variance explained by the model, with $\mathrm{NSE}=1$ indicating perfect prediction and $\mathrm{NSE}=0$ equivalent to using the historical mean as a constant forecast. Because wind power time series exhibit large variance driven by rapid wind fluctuations, the denominator in NSE is inherently large; consequently, NSE values in wind power forecasting are typically lower than those reported in hydrology or other domains with smoother signals. We include NSE as a unit-normalized complement to MSE and MAE for interpretability.

\subsubsection{Implementation Details}
\label{sec:impl}

All experiments are implemented in PyTorch and conducted on a single NVIDIA GPU. The input sequence length is set to $T{=}96$ (24 hours at 15-minute resolution) and the prediction horizon to $H{=}16$ (4 hours). Table~\ref{tab:hyperparams} lists the key hyperparameters of ECTO.

\begin{table}[htbp]
\centering
\caption{Key hyperparameters of ECTO. All submodule hidden dimensions ($d$, $d_e$, $d_s$) are set to $d_{\mathrm{model}}$ unless otherwise noted.}
\label{tab:hyperparams}
\begin{tabular}{ll}
\toprule
Hyperparameter & Value \\
\midrule
Input length $T$ & 96 \\
Prediction horizon $H$ & 16 \\
Model dimension $d_{\mathrm{model}}$ & 256 \\
Local conv. channels $c$ & 32 \\
Token kernel / stride & 16 / 8 \\
Number of regime slots $K$ & 4 \\
Short-window size $T_s$ & 12 \\
Long-window size $T_l$ & 48 \\
Group temperature $\tau_g$ & 1.0 (fixed) \\
Variable temperature $\tau_v$ & 1.0 (fixed) \\
Top-$k$ (variable selection) & 2 \\
MAE weight $\lambda$ & 0.05 \\
Embedding dropout & 0.1 \\
Calibration dropout & 0.1 \\
Prediction head dropout & 0.3 \\
Batch size & 256 \\
Learning rate (peak) & $2 \times 10^{-4}$ \\
Optimizer & Adam \\
LR scheduler & OneCycleLR (cosine, pct\_start=0.3) \\
Max epochs & 50 \\
Early stopping patience & 10 (on validation loss) \\
\bottomrule
\end{tabular}
\end{table}

For all baselines, we use the same input/output lengths, batch size, learning rate, and early stopping patience to ensure a fair comparison. Each model is tuned following the recommendations in its original paper. Multi-seed statistics are reported separately where indicated.

\subsection{Main Results}
\label{sec:main-results}

Table~\ref{tab:main-results} presents the forecasting results on all three datasets. The best results are boldfaced and the second-best are underlined. All values are reported using seed 2025; Table~\ref{tab:multiseed} provides multi-seed verification (mean$\pm$std over three seeds) on WF1.

\begin{table}[htbp]
\centering
\caption{Forecasting results on three wind farms (seed=2025, $H{=}16$). The best results are boldfaced and the second-best are underlined. RMSE is reported in MW to facilitate practical interpretation.}
\label{tab:main-results}
\resizebox{\textwidth}{!}{
\begin{tabular}{lcccccccccccc}
\toprule
& \multicolumn{4}{c}{WF1 (99\,MW)} & \multicolumn{4}{c}{WF4 (66\,MW)} & \multicolumn{4}{c}{Xinjiang (200\,MW)} \\
\cmidrule(lr){2-5} \cmidrule(lr){6-9} \cmidrule(lr){10-13}
Model & MSE & MAE & NSE & RMSE & MSE & MAE & NSE & RMSE & MSE & MAE & NSE & RMSE \\
& & & & (MW) & & & & (MW) & & & & (MW) \\
\midrule
Persistence & 0.427 & 0.382 & 0.648 & 15.22 & 0.144 & 0.210 & 0.867 & 7.53 & 0.308 & 0.305 & 0.684 & 36.74 \\
Informer & 0.377 & 0.402 & 0.689 & 14.29 & 0.156 & 0.253 & 0.856 & 7.85 & 0.353 & 0.391 & 0.637 & 39.38 \\
DLinear & 0.402 & 0.427 & 0.668 & 14.77 & 0.167 & 0.276 & 0.846 & 8.11 & 0.365 & 0.453 & 0.625 & 40.03 \\
PatchTST & 0.387 & 0.396 & 0.681 & 14.49 & 0.140 & 0.223 & 0.871 & 7.43 & \underline{0.285} & \underline{0.340} & \textbf{0.708} & 35.34 \\
TimesNet & 0.403 & 0.400 & 0.668 & 14.77 & 0.156 & 0.242 & 0.856 & 7.84 & 0.337 & 0.365 & 0.654 & 38.48 \\
iTransformer & 0.399 & 0.389 & 0.671 & 14.71 & 0.146 & 0.227 & 0.865 & 7.60 & 0.317 & 0.352 & 0.675 & 37.29 \\
TimeMixer & 0.400 & 0.396 & 0.671 & 14.72 & 0.147 & 0.223 & 0.864 & 7.62 & 0.325 & 0.353 & 0.666 & 37.76 \\
CrossLinear & 0.385 & 0.393 & 0.683 & 14.44 & 0.142 & 0.229 & 0.869 & 7.50 & 0.311 & 0.351 & 0.681 & 36.92 \\
TimeXer & 0.387 & 0.391 & 0.681 & 14.48 & 0.138 & 0.227 & 0.873 & 7.38 & 0.298 & 0.337 & 0.694 & 36.14 \\
XLinear & \underline{0.380} & \textbf{0.379} & \underline{0.689} & 14.35 & \underline{0.134} & \textbf{0.212} & \underline{0.877} & 7.27 & 0.288 & \underline{0.327} & 0.691 & 35.57 \\
\midrule
\textbf{ECTO} & \textbf{0.366} & \underline{0.386} & \textbf{0.701} & \textbf{14.08} & \textbf{0.131} & \underline{0.217} & \textbf{0.879} & \textbf{7.20} & \textbf{0.273} & \textbf{0.321} & \underline{0.707} & \textbf{34.62} \\
\bottomrule
\end{tabular}
}
\end{table}

\begin{table}[htbp]
\centering
\caption{Multi-seed verification on WF1 (mean$\pm$std over seeds 2024, 2025, 2026). WF1 is chosen as the representative site for multi-seed analysis because it has the most balanced power distribution (zero-power ratio 0.29\%) and moderate exogenous dimensionality, making it the most representative test of training stability. Results on WF4 and Xinjiang are provided in Table~\ref{tab:main-results} with seed~2025. The rank order of models is consistent across all three datasets, suggesting that the relative conclusions generalize.}
\label{tab:multiseed}
\begin{tabular}{lccc}
\toprule
Model & MSE $\downarrow$ & MAE $\downarrow$ & NSE $\uparrow$ \\
\midrule
Informer & $0.377_{\pm0.001}$ & $0.402_{\pm0.000}$ & $0.689_{\pm0.000}$ \\
XLinear & $0.380_{\pm0.000}$ & $\mathbf{0.378}_{\pm0.001}$ & $0.689_{\pm0.000}$ \\
CrossLinear & $0.385_{\pm0.001}$ & $0.391_{\pm0.002}$ & $0.683_{\pm0.001}$ \\
TimeXer & $0.386_{\pm0.001}$ & $0.388_{\pm0.003}$ & $0.682_{\pm0.001}$ \\
PatchTST & $0.387_{\pm0.000}$ & $0.395_{\pm0.001}$ & $0.681_{\pm0.000}$ \\
iTransformer & $0.397_{\pm0.001}$ & $0.388_{\pm0.001}$ & $0.672_{\pm0.001}$ \\
TimeMixer & $0.397_{\pm0.004}$ & $0.392_{\pm0.003}$ & $0.673_{\pm0.003}$ \\
DLinear & $0.402_{\pm0.000}$ & $0.427_{\pm0.000}$ & $0.668_{\pm0.000}$ \\
TimesNet & $0.410_{\pm0.006}$ & $0.405_{\pm0.005}$ & $0.662_{\pm0.005}$ \\
\midrule
\textbf{ECTO} & $\mathbf{0.366}_{\pm0.001}$ & $0.380_{\pm0.005}$ & $\mathbf{0.700}_{\pm0.001}$ \\
\bottomrule
\end{tabular}
\end{table}

Several observations can be drawn from the results.

\textbf{Persistence reference.} The persistence baseline---using the last observed value as the forecast for all horizons---yields MSE of 0.427 (WF1), 0.144 (WF4), and 0.308 (Xinjiang) at $H{=}16$. On WF4 and Xinjiang, persistence outperforms more than half of the deep learning baselines, underscoring that generic time series models without proper exogenous handling can fail to improve over this simple reference. In contrast, ECTO achieves 14.3\%, 9.0\%, and 11.4\% lower MSE than persistence on the three sites, respectively. At $H{=}32$ (Table~\ref{tab:multi-horizon}), persistence degrades sharply (e.g., NSE drops from 0.648 to 0.435 on WF1), while ECTO's advantage widens to 17--29\%, confirming that learned representations become increasingly valuable as forecast uncertainty grows.

\textbf{Overall performance.} ECTO achieves the lowest MSE across all three datasets, with relative improvements over the strongest baseline (XLinear) ranging from 2.2\% on WF4 to 5.2\% on Xinjiang. This advantage extends to the absolute error in MW: ECTO attains the lowest RMSE on all three sites---14.08\,MW on WF1, 7.20\,MW on WF4, and 34.62\,MW on Xinjiang (Table~\ref{tab:main-results})---corresponding to rated-capacity-normalized RMSE of 14.2\%, 10.9\%, and 17.3\%, respectively. ECTO also exhibits the smallest systematic bias (MBE) on WF1 ($-$0.03\,MW) and WF4 (0.06\,MW), indicating that its predictions are not only more accurate but also better calibrated against persistent over- or under-prediction. Among all baselines published before 2026, ECTO achieves the best MSE and MAE simultaneously; it is only with the inclusion of XLinear (2026) that a marginal MAE gap emerges on WF1 and WF4 (0.007 and 0.005, respectively). Notably, ECTO attains both the lowest MSE and MAE on Xinjiang, the site with the largest capacity and highest-dimensional exogenous space. The MSE--MAE trade-off on WF1 and WF4 reflects a systematic difference in error distributions rather than a random fluctuation: MSE penalizes large errors quadratically, whereas MAE weights all errors uniformly. The error decomposition in Table~\ref{tab:error-decomp} confirms that ECTO achieves the lowest MSE during ramp events and in the 20--80\% power range---precisely the regimes where large, dispatch-critical errors concentrate---while XLinear achieves lower MAE in stable and near-rated conditions, where errors are small and uniformly distributed. In other words, ECTO reduces the error tail at the cost of a slight increase in average error, a trade-off that is favorable for grid dispatch applications where avoiding large prediction failures matters more than minimizing average deviation. On WF1 and WF4, ECTO also attains the highest NSE (0.701 and 0.879, respectively). On Xinjiang, ECTO's NSE (0.707) is marginally below PatchTST (0.708), despite having a 4.2\% lower MSE; this is because NSE and MSE weight individual errors differently, and the two models differ mainly in how they handle a small subset of extreme samples. Diebold-Mariano tests with Newey-West HAC correction (Bartlett kernel, bandwidth $h=H=16$) confirm that ECTO's MSE advantage over XLinear is statistically significant across all three wind farms (WF1: DM~$\operatorname{=}-7.6$, $p\,{<}\,0.001$; WF4: DM~$\operatorname{=}-3.2$, $p\,{<}\,0.001$; Xinjiang: DM~$\operatorname{=}-7.2$, $p\,{<}\,0.001$). As shown in Table~\ref{tab:multiseed}, ECTO's advantage is consistent across random seeds with low variance.

\textbf{Value of exogenous variable modeling.} Models that explicitly incorporate exogenous information (TimeXer, XLinear, and ECTO) generally outperform target-only architectures. However, TimeXer, which uses cross-attention to inject exogenous information without variable selection, achieves comparable or worse results than PatchTST on WF4 and Xinjiang. This suggests that simply providing exogenous variables to the model is insufficient; the key lies in identifying \emph{which} variables are informative under the current conditions---precisely the role of PGVS.

\textbf{Attention vs.\ linear architectures.} The lightweight DLinear model, despite having far fewer parameters, remains competitive on WF4 (NSE = 0.846), underscoring that short-horizon wind power forecasting exhibits strong local linearity. Nevertheless, ECTO's advantage grows on Xinjiang (+5.2\% MSE over XLinear), where the higher-dimensional exogenous space (13 variables) and larger capacity factor (35.1\%) reward a more structured approach to exogenous variable selection and conditioning.

\textbf{Multi-horizon generalization.} Table~\ref{tab:multi-horizon} extends the comparison to a longer prediction horizon ($H{=}32$, 8 hours ahead). ECTO retains the lowest MSE across both sites and both horizons, confirming that its exogenous conditioning mechanism remains effective beyond the default 4-hour setting. On WF1, ECTO's advantage over XLinear widens from 3.7\% at $H{=}16$ to 8.6\% at $H{=}32$, suggesting that the horizon-specific refinement in ECRR (Section~\ref{sec:horizon}) becomes increasingly valuable as forecast uncertainty grows with prediction length. This widening margin is consistent with the per-step trend observed in Fig.~\ref{fig:horizon-mse} and supports the hypothesis that the ECRR horizon embedding learns to calibrate systematically across different lead times. On Xinjiang, ECTO remains the strongest model at $H{=}32$ (MSE~0.476), though the gap over PatchTST narrows, reflecting the different error growth patterns in the arid northwestern climate. Across both sites and horizons, DLinear and CrossLinear are the closest competitors, while Informer exhibits the largest degradation at the longer horizon.

\begin{table}[htbp]
\centering
\caption{Multi-horizon forecasting results (MSE) on WF1 and Xinjiang. $H{=}16$ corresponds to 4-hour ahead, $H{=}32$ to 8-hour ahead. All results are reported with seed 2025. The best results are boldfaced.}
\label{tab:multi-horizon}
\begin{tabular}{lcccc}
\toprule
& \multicolumn{2}{c}{WF1} & \multicolumn{2}{c}{Xinjiang} \\
\cmidrule(lr){2-3} \cmidrule(lr){4-5}
Model & $H{=}16$ & $H{=}32$ & $H{=}16$ & $H{=}32$ \\
\midrule
Persistence  & 0.427 & 0.686 & 0.308 & 0.572 \\
\textbf{ECTO}    & \textbf{0.366} & \textbf{0.568} & \textbf{0.273} & \textbf{0.476} \\
XLinear     & 0.380 & 0.621 & 0.288 & 0.497 \\
PatchTST    & 0.387 & 0.588 & 0.285 & 0.485 \\
CrossLinear & 0.385 & 0.589 & 0.311 & 0.540 \\
TimeXer     & 0.387 & 0.588 & 0.298 & 0.507 \\
DLinear     & 0.402 & 0.579 & 0.365 & 0.536 \\
Informer    & 0.377 & 0.635 & 0.353 & 0.565 \\
\bottomrule
\end{tabular}
\end{table}

Table~\ref{tab:efficiency} compares the computational efficiency of all models on WF1. ECTO has 0.93\,M parameters, placing it in the mid-range---smaller than CrossLinear (3.73\,M), TimeXer (5.38\,M), and comparable to TimesNet (1.18\,M). Its average epoch time (4.10\,s) is significantly faster than Transformer-based models with complex architectures (TimesNet 48.70\,s, TimeMixer 13.13\,s, Informer 11.95\,s, PatchTST 11.89\,s) and comparable to CrossLinear (3.58\,s) and TimeXer (4.33\,s), while maintaining a low peak GPU memory footprint (0.11\,GB). These results confirm that ECTO's accuracy improvements do not come at prohibitive computational cost.

\begin{table}[htbp]
\centering
\caption{Computational efficiency on WF1 ($T{=}96$, $H{=}16$, batch size 256, single GPU). Params: trainable parameters. Avg.\ Epoch: mean training-loop time per epoch (excl.\ validation). Peak Mem: peak GPU memory allocated by PyTorch. Sorted by Avg.\ Epoch.}
\label{tab:efficiency}
\begin{tabular}{lccc}
\toprule
Model & Params (M) & Avg.\ Epoch (s) & Peak Mem (GB) \\
\midrule
DLinear & 0.003 & 1.43 & 0.02 \\
XLinear & 0.593 & 1.47 & 0.08 \\
iTransformer & 0.214 & 2.62 & 0.10 \\
CrossLinear & 3.729 & 3.58 & 0.10 \\
\textbf{ECTO} & \textbf{0.932} & \textbf{4.10} & \textbf{0.11} \\
TimeXer & 5.376 & 4.33 & 0.24 \\
Informer & 0.427 & 11.95 & 0.57 \\
PatchTST & 0.326 & 11.89 & 0.88 \\
TimeMixer & 0.090 & 13.13 & 1.26 \\
TimesNet & 1.184 & 48.70 & 0.76 \\
\bottomrule
\end{tabular}
\end{table}

\subsubsection{Prediction Curves}

\begin{figure}[htbp]
\centering
\includegraphics[width=\textwidth]{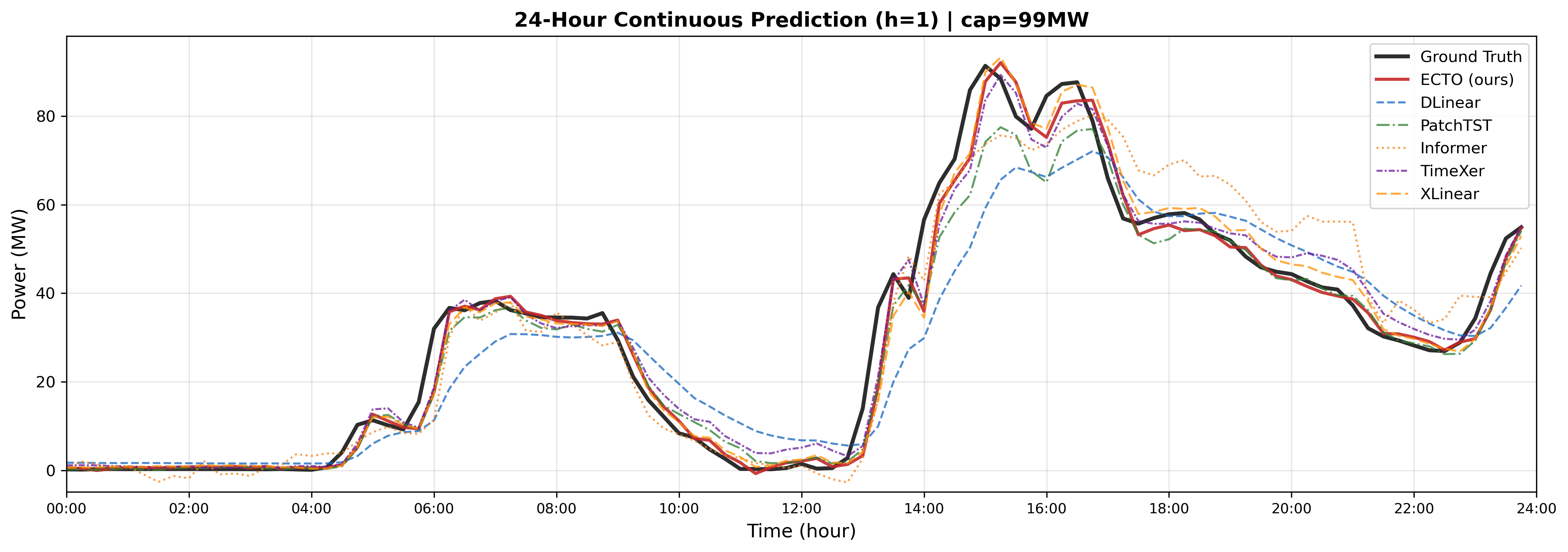}
\caption{24-hour continuous prediction on WF1 (day starting at sample 8976). Each point is the 15-minute-ahead forecast (horizon $h=1$). The day contains 24 ramp events ($\geq 5$\,MW change), providing a demanding test of ECTO's behavior under frequent power transitions.}
\label{fig:daily-pred}
\end{figure}

To provide qualitative insight into forecast behavior, Fig.~\ref{fig:daily-pred} shows a 24-hour continuous prediction curve on WF1, comparing ECTO against DLinear, PatchTST, and Informer. The selected day spans nearly the full power range (0.1--91.4\,MW) and contains 24 ramp events, making it a demanding test of model behavior under frequent transitions. ECTO tracks the ground truth more faithfully than the three baselines throughout the day: during the steep ramp-down at hour 10 and the multi-peak fluctuation between hours 15--20, ECTO exhibits less overshoot and recovers more quickly, while DLinear and Informer produce larger transient deviations. This visual behavior is consistent with the MSE advantage in Table~\ref{tab:main-results} and the error decomposition (Section~\ref{sec:error-decomp}), where ECTO outperforms baselines most strongly on ramp events.

\begin{figure}[htbp]
\centering
\begin{subfigure}[b]{0.48\textwidth}
\centering
\includegraphics[width=\textwidth]{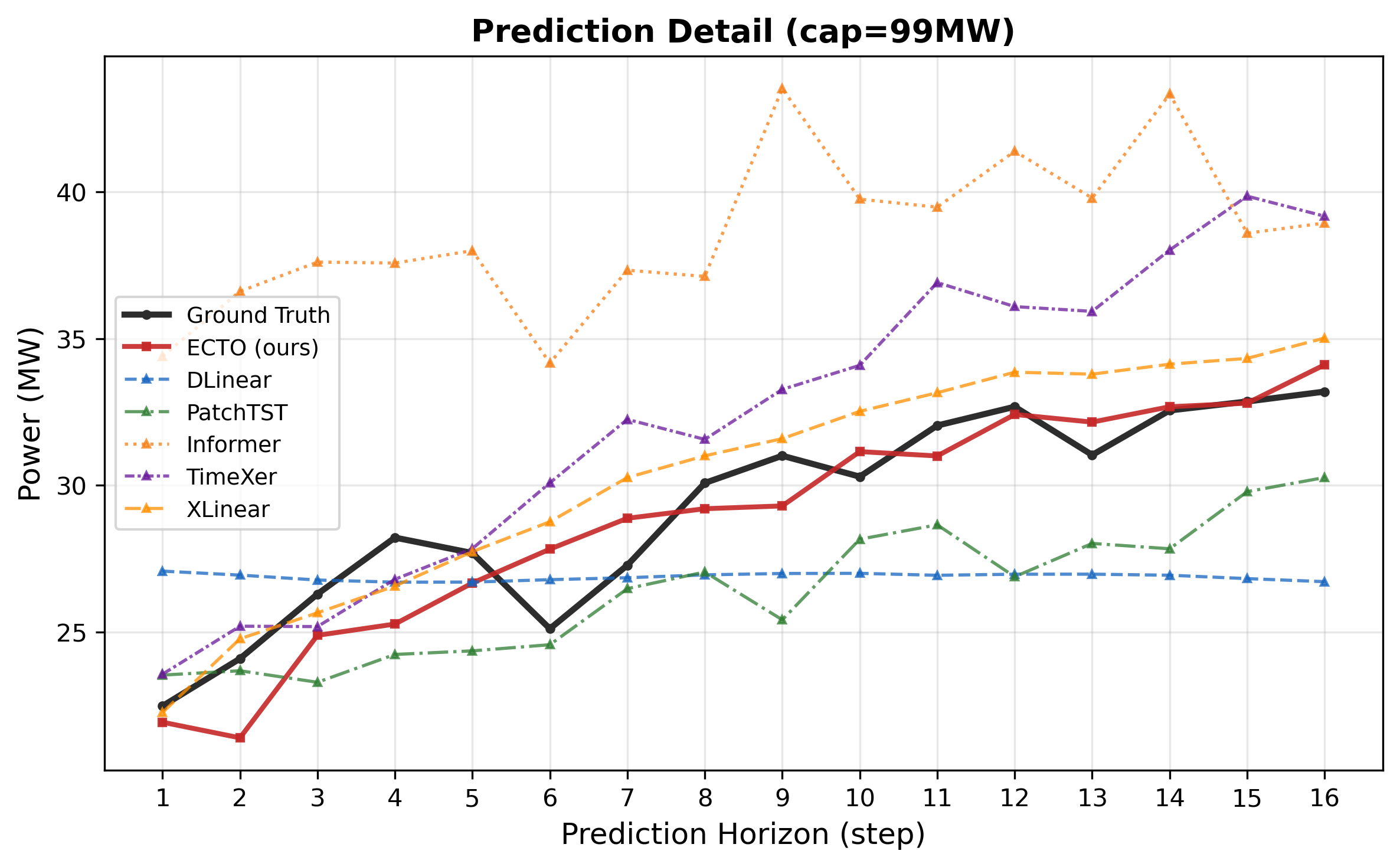}
\caption{Ramp Up}
\label{fig:pred-rampup}
\end{subfigure}
\hfill
\begin{subfigure}[b]{0.48\textwidth}
\centering
\includegraphics[width=\textwidth]{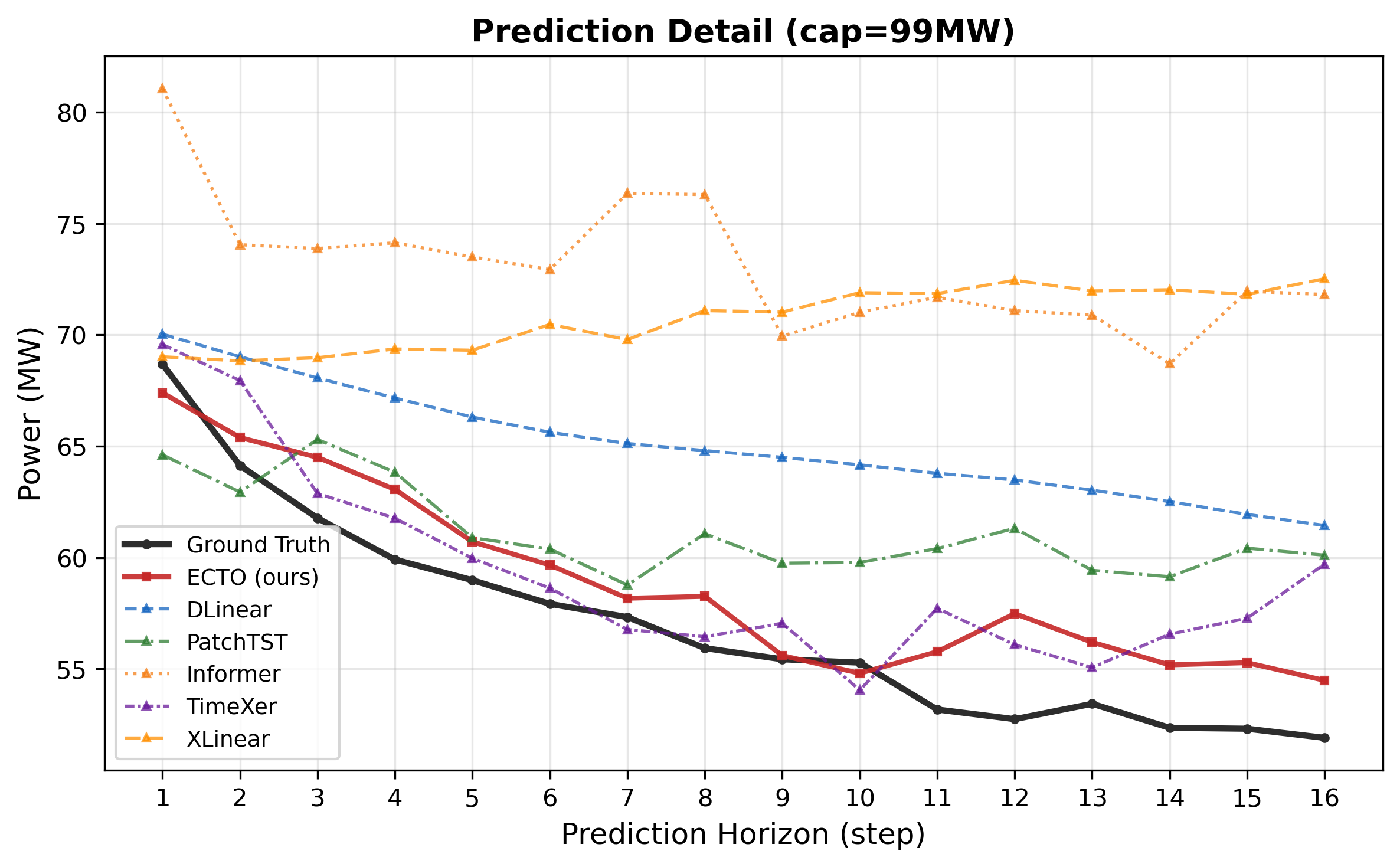}
\caption{Ramp Down}
\label{fig:pred-rampdown}
\end{subfigure}

\vspace{6pt}
\begin{subfigure}[b]{0.48\textwidth}
\centering
\includegraphics[width=\textwidth]{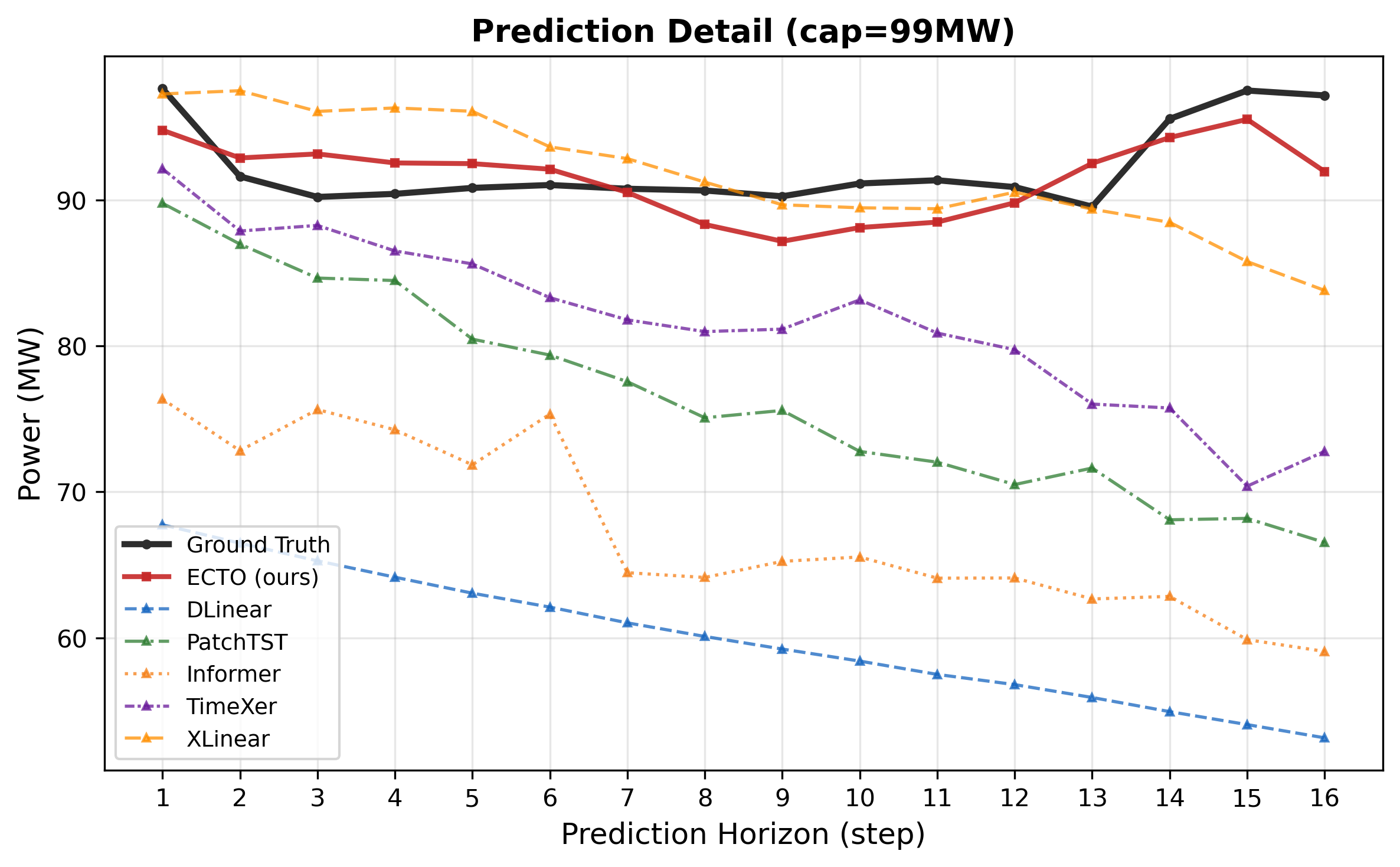}
\caption{High Power}
\label{fig:pred-highpower}
\end{subfigure}
\hfill
\begin{subfigure}[b]{0.48\textwidth}
\centering
\includegraphics[width=\textwidth]{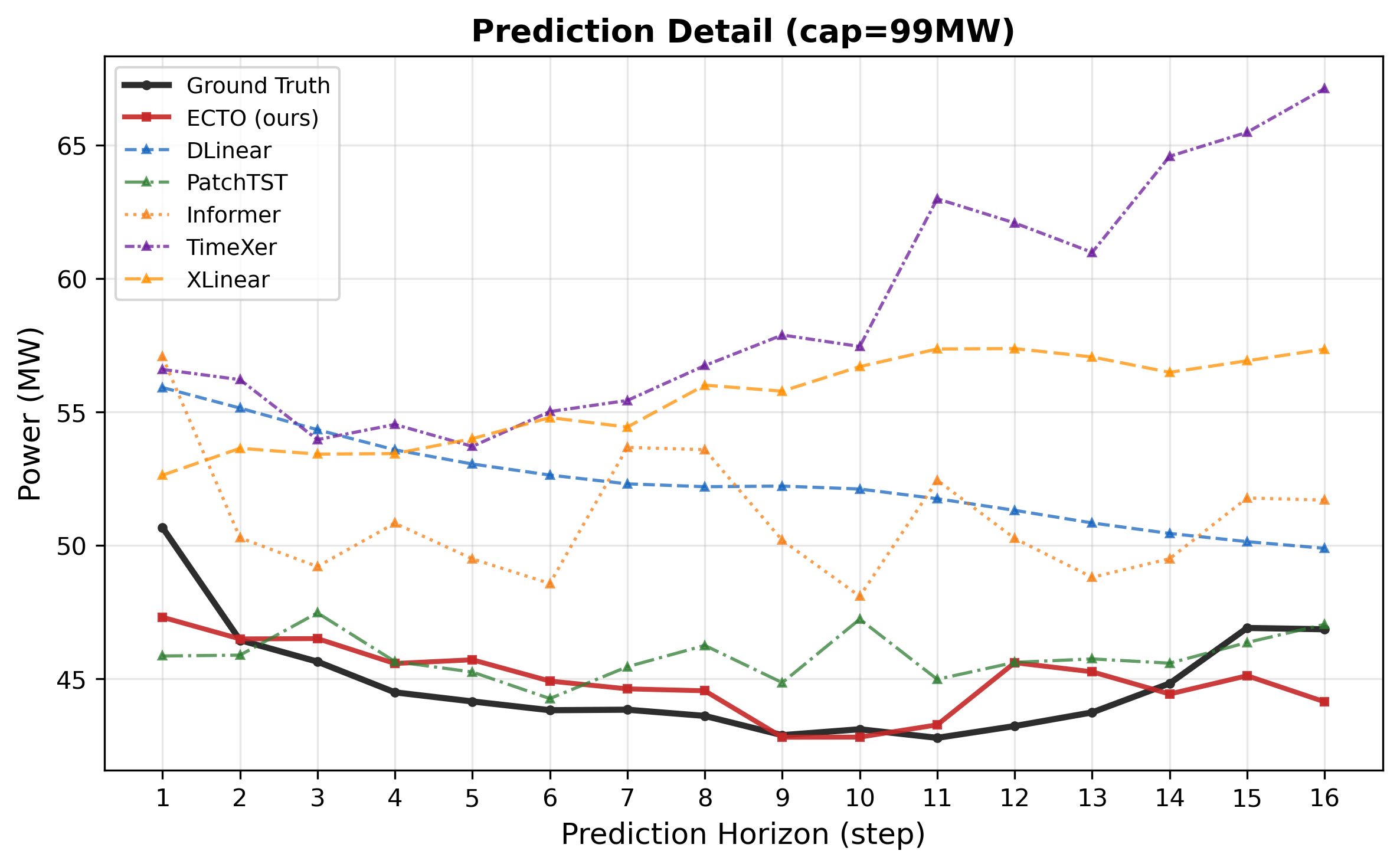}
\caption{Medium Stable}
\label{fig:pred-stable}
\end{subfigure}
\caption{16-step prediction details across four representative operating conditions on WF1. Each panel shows the 4-hour forecast window ($h=1$ to $h=16$); the ground truth is in black. ECTO is compared against DLinear, PatchTST, and Informer.}
\label{fig:pred-scenes}
\end{figure}

Fig.~\ref{fig:pred-scenes} zooms into four qualitatively distinct operating conditions. In the Ramp Up and Ramp Down panels, ECTO tracks the monotonic trend with less oscillatory deviation than the baselines, particularly Informer which tends to overreact to local fluctuations. In the High Power panel (93\% rated), where the power curve saturates, all models produce tighter predictions, but ECTO stays closest to the truth across all 16 steps. In the Medium Stable panel, all models perform comparably, confirming that the primary source of ECTO's advantage lies in dynamic rather than quiescent conditions.

\begin{figure}[htbp]
\centering
\includegraphics[width=0.65\textwidth]{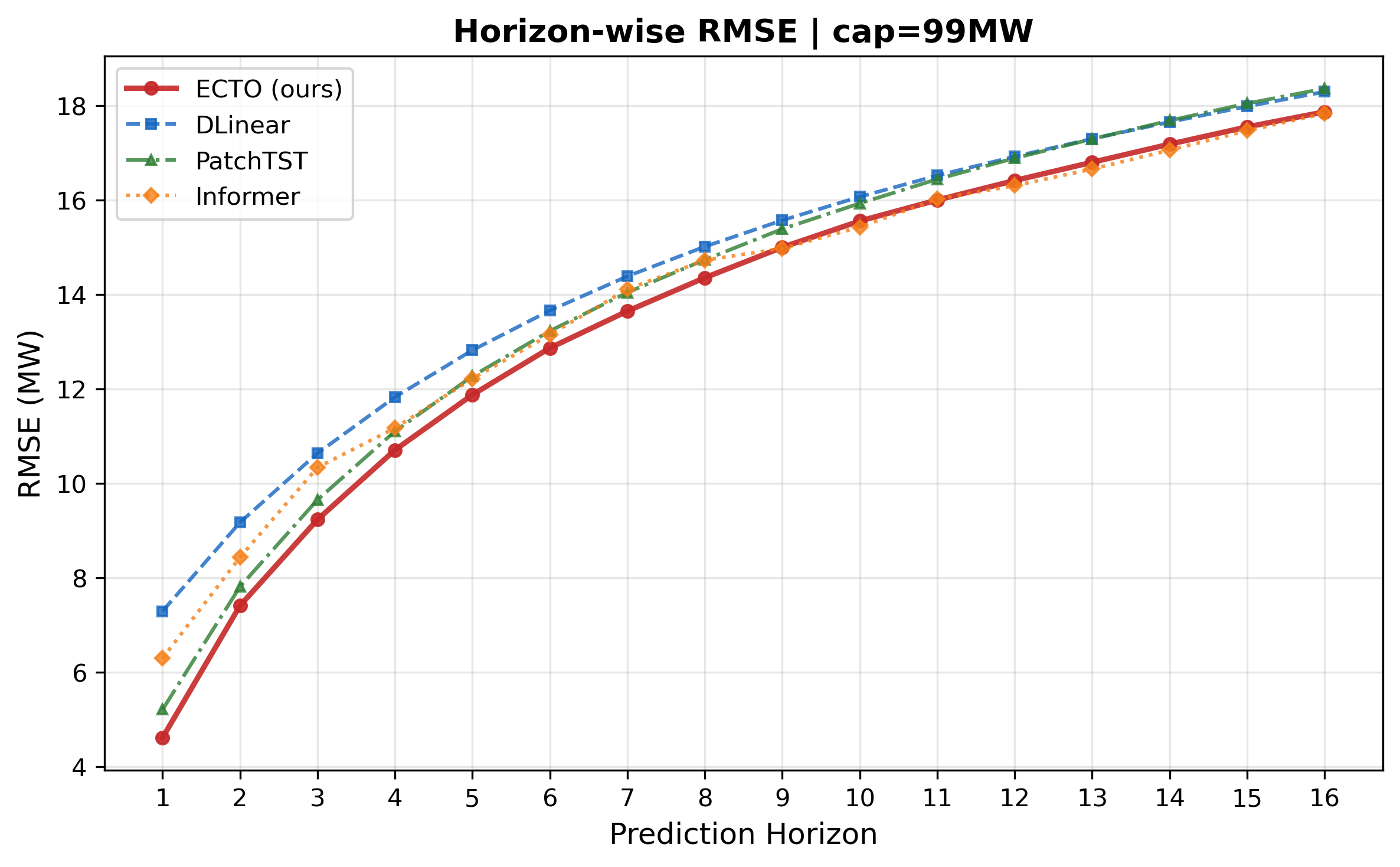}
\caption{Horizon-wise RMSE on the full WF1 test set. Each point is the RMSE at a specific forecast horizon aggregated over all 13,824 test samples.}
\label{fig:horizon-rmse}
\end{figure}

To verify that the above examples are representative rather than cherry-picked, Fig.~\ref{fig:horizon-rmse} aggregates the horizon-wise RMSE over the entire WF1 test set. ECTO achieves the lowest RMSE at all horizons, with the gap being largest in the short-to-medium range ($h=1$--8). Beyond $h=8$, all models converge as forecast uncertainty grows, which is expected for ultra-short-term wind forecasting. The consistent full-test-set advantage across all 16 horizons confirms that ECTO's improvements are systematic.

\begin{figure}[htbp]
\centering
\includegraphics[width=\textwidth]{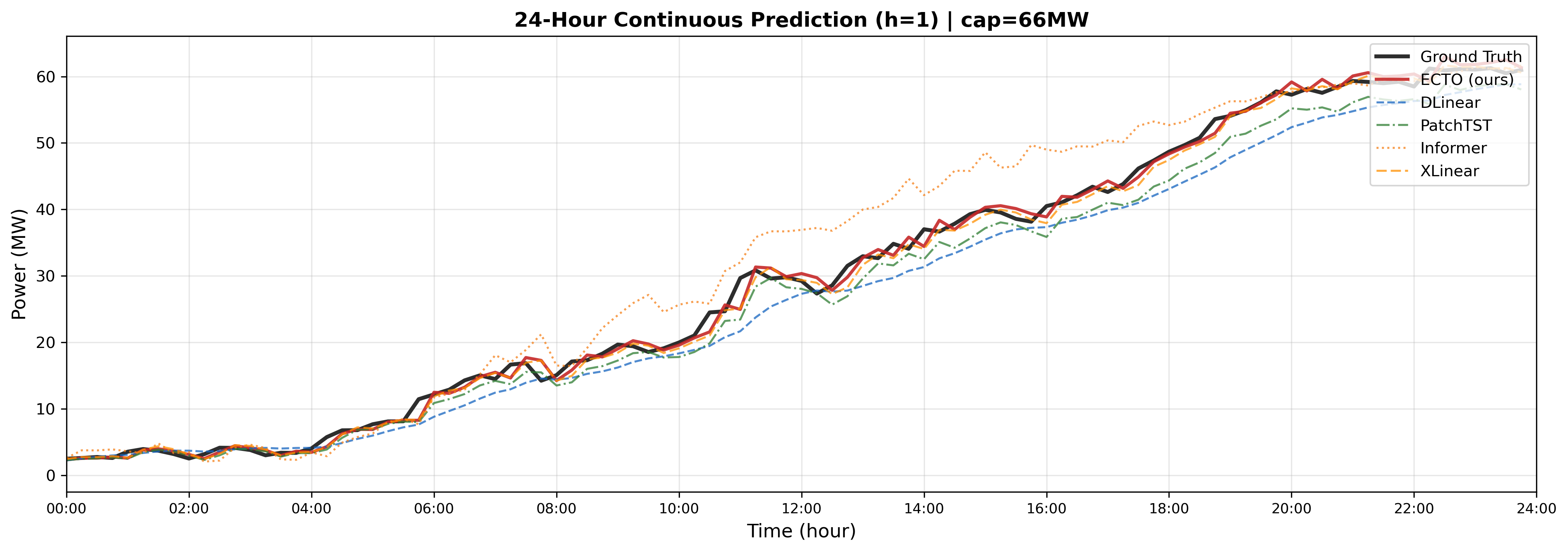}
\caption{24-hour continuous prediction on WF4 (66\,MW, day starting at sample 9696).}
\label{fig:daily-pred-wf4}
\end{figure}

To assess cross-site generalization of the qualitative behavior, Fig.~\ref{fig:daily-pred-wf4} shows a 24-hour continuous prediction on WF4, which has a substantially different power distribution from WF1 (19.69\% zero-power ratio vs.\ 0.29\%). The selected day captures a transition from low wind to strong generation, with ECTO tracking the steep power ramp more faithfully than the baselines and exhibiting less overshoot during the rapid ascent.

\subsection{Ablation Study}
\label{sec:ablation}

To evaluate the contribution of each proposed component, we conduct an ablation study on WF1 using seed 2025. Table~\ref{tab:ablation} reports the results. All variants share the same backbone (TTE) and training protocol; only the specified module is removed or modified.

\begin{table}[t]
\centering
\caption{Ablation study on WF1 (seed 2025). $\Delta$MSE denotes the relative increase over Full ECTO.}
\label{tab:ablation}
\begin{tabular}{llcc}
\toprule
Module & Variant & MSE & $\Delta$MSE \\
\midrule
--- & Target-only (no exogenous) & 0.3826 & +4.59\% \\
\midrule
\multirow{4}{*}{PGVS} & 1-group Variant & 0.3711 & +1.47\% \\
 & w/o PGVS & 0.3725 & +1.84\% \\
 & w/o Group Scoring & 0.3719 & +1.68\% \\
 & w/o TopK Selection & 0.3706 & +1.32\% \\
\midrule
\multirow{3}{*}{ECRR} & w/o ECRR & 0.3763 & +2.86\% \\
 & w/o Regime Refinement & 0.3714 & +1.53\% \\
 & w/o Horizon Refinement & 0.3711 & +1.44\% \\
\midrule
\textbf{---} & \textbf{Full ECTO} & \textbf{0.3658} & --- \\
\bottomrule
\end{tabular}
\end{table}

Several observations can be drawn.

\textbf{Value of exogenous variables.} Removing all exogenous conditioning (Target-only) degrades MSE by 4.59\%, confirming that meteorological and wind-speed variables carry substantial predictive information beyond what is encoded in the target power sequence alone.

\textbf{Physical grouping and variable selection.} The 1-group Variant collapses the two physically motivated groups (wind speed and meteorological) into a single flat group, disabling the hierarchical selection structure. This increases MSE by 1.47\%, demonstrating that the physical grouping prior effectively constrains the variable search space. Within PGVS, removing the group scoring mechanism (uniform group weights) degrades MSE by 1.68\%, and removing the top-$k$ variable selection within each active group yields a degradation of 1.32\%, indicating that fine-grained sparsity provides a consistent benefit. Removing PGVS entirely---i.e., disabling both physical grouping, group scoring, and top-$k$ selection, so that all exogenous variables contribute uniformly to the downstream modules---degrades MSE by 1.84\%. These results confirm that the three mechanisms complement each other: physical grouping provides the structural scaffold, group scoring allocates attention across groups, and top-$k$ selection prunes within each group.

\textbf{Exogenous-Conditioned Regime Refinement.} Removing the entire ECRR module degrades MSE by 2.86\%, making it the single largest contributor among ECTO's exogenous-related components. Decomposing ECRR into its two sub-modules reveals that regime refinement (+1.53\%) and horizon refinement (+1.44\%) contribute roughly equally, indicating that both operating-condition routing and horizon-specific calibration are important for multi-step wind power forecasting. The sum of the two sub-module degradations (2.97\%) closely matches the full ECRR degradation (2.86\%), confirming that the two mechanisms operate largely independently. This confirms that the exogenous context produced by PGVS ($\mathbf{z}^{\mathrm{exo}}$) is effectively consumed by ECRR to produce regime-aware, horizon-specific calibration, forming the primary pathway through which exogenous information influences the final prediction.

\textbf{Multi-seed consistency.} To verify that the above conclusions are not seed-dependent, we repeat the ablation with three random seeds (2024, 2025, 2026). The rank order of degradation is consistent across seeds: physical grouping (+1.79\% mean $\Delta$MSE), group scoring (+1.28\%), top-$k$ selection (+0.97\%), and the full ECRR (+2.08\%) all show stable positive degradation when removed.

\textbf{Hyperparameter sensitivity.} Table~\ref{tab:sensitivity} reports the sensitivity of ECTO to the two key structural hyperparameters: the number of physical groups $G$ and the number of regime experts $K$. For the grouping strategy, $G{=}2$ yields the lowest MSE, outperforming both $G{=}1$ (+2.2\%) and $G{=}3$ (+1.9\%). The degradation of $G{=}3$ arises because wind direction is a $360^\circ$ cyclic variable whose group-level average carries limited discriminative power; placing direction variables alongside temperature, pressure, and humidity in a single atmospheric group allows the cross-attention mechanism to extract marginal directional information without aggregating it into a meaningless summary token. For the regime count, $K{=}3$ and $K{=}4$ lie in a flat optimal basin with nearly identical MSE (0.3657 vs.\ 0.3658), while $K{=}2$ underfits (+1.9\%) and $K{=}5$ begins to fragment the training samples (+1.0\%). We adopt $K{=}4$ because it incurs no accuracy penalty over $K{=}3$ while providing additional capacity for regime discrimination; notably, the interpretability analysis (Section~\ref{sec:interpret}) shows that even with $K{=}4$ available experts, ECRR converges to only two well-separated calibration strategies across all sites, confirming that the router naturally sparsifies its expert usage.

\begin{table}[htbp]
\centering
\caption{Hyperparameter sensitivity on WF1 (seed 2025). $\Delta$MSE is relative to the default configuration ($G{=}2$, $K{=}4$).}
\label{tab:sensitivity}
\begin{tabular}{lcclcc}
\toprule
\multicolumn{3}{c}{\textbf{Grouping strategy} ($K{=}4$)} & \phantom{aa} & \multicolumn{2}{c}{\textbf{Regime count} ($G{=}2$)} \\
\cmidrule(lr){1-3} \cmidrule(lr){5-6}
$G$ & Configuration & MSE ($\Delta$) & & $K$ & MSE ($\Delta$) \\
\midrule
1 & Wind speed only & 0.374 (+2.2\%) & & 2 & 0.372 (+1.9\%) \\
2 & Wind + Atm.\ (no dir.) & 0.369 (+1.0\%) & & 3 & 0.366 ({\textasciitilde}0\%) \\
2 & Wind + Atm.\ (ours) & \textbf{0.366} & & \textbf{4} & \textbf{0.366} \\
3 & Wind // Meteo // Dir. & 0.373 (+1.9\%) & & 5 & 0.369 (+1.0\%) \\
\bottomrule
\end{tabular}
\end{table}

\subsection{Error Decomposition Analysis}
\label{sec:error-decomp}

To understand \emph{where} ECTO's MSE advantage originates, we decompose the prediction errors on WF1 along three dimensions: power level, ramp intensity, and prediction horizon. This analysis reveals whether the overall improvement is uniformly distributed or concentrated in specific operating regimes.

\begin{table}[htbp]
\centering
\caption{Error decomposition on WF1 (MSE in MW$^2$). Test samples are segmented by power level (top) and ramp intensity (bottom). The best results are boldfaced.}
\label{tab:error-decomp}
\resizebox{\textwidth}{!}{
\begin{tabular}{lcccccccc}
\toprule
& ECTO & PatchTST & TimesNet & iTransformer & TimeMixer & CrossLinear & TimeXer & XLinear \\
\midrule
\multicolumn{9}{l}{\textit{Power level (\% rated capacity)}} \\
0--5\% & 33.7 & 32.0 & 28.6 & \textbf{25.5} & 28.5 & 34.7 & 35.4 & 24.9 \\
5--20\% & 122.0 & 122.0 & \textbf{109.7} & 116.2 & 124.0 & 119.4 & 122.9 & 116.6 \\
20--50\% & \textbf{324.9} & 351.2 & 363.3 & 374.6 & 375.5 & 357.5 & 357.6 & 362.9 \\
50--80\% & \textbf{440.6} & 462.1 & 514.2 & 493.9 & 474.0 & 461.5 & 448.8 & 462.7 \\
80--100\% & 426.4 & 576.9 & 619.5 & 544.5 & 555.1 & 431.4 & 435.6 & \textbf{388.8} \\
\midrule
\multicolumn{9}{l}{\textit{Ramp intensity}} \\
Ramp Up & \textbf{740.0} & 781.0 & 825.9 & 831.1 & 807.7 & 770.4 & 746.9 & 792.1 \\
Ramp Down & \textbf{412.2} & 444.8 & 430.3 & 461.4 & 491.8 & 459.2 & 474.6 & 464.6 \\
Stable & 103.8 & 110.3 & 116.3 & 109.9 & 109.3 & 107.9 & 108.0 & \textbf{100.3} \\
\bottomrule
\end{tabular}
}
\end{table}

\textbf{ECTO excels in challenging regimes.} Table~\ref{tab:error-decomp} reveals a clear pattern: ECTO achieves the lowest MSE in the mid-power range (20--50\% and 50--80\% of rated capacity) and during both ramp-up and ramp-down events. These are precisely the operating conditions where forecasting errors have the greatest impact on grid dispatch decisions. In the 20--50\% range, ECTO's MSE (324.9) is 7.5\% lower than the next-best CrossLinear (357.5); during ramp-down events, the gap widens to 4.2\% over TimesNet (430.3). This concentration of improvement in high-error regimes explains ECTO's overall MSE advantage despite not being uniformly superior across all conditions.

\textbf{XLinear excels in steady-state and extreme conditions.} XLinear attains the lowest MSE in the 80--100\% bin (388.8 vs.\ ECTO's 426.4) and during stable periods (100.3 vs.\ ECTO's 103.8). Its variable-wise sigmoid gating provides a smooth, continuous modulation that performs well when the power signal is either near its ceiling or evolving slowly. However, this advantage diminishes in the mid-power range and vanishes during ramp events, where abrupt transitions demand more dynamic exogenous conditioning.

\begin{figure}[htbp]
\centering
\includegraphics[width=\textwidth]{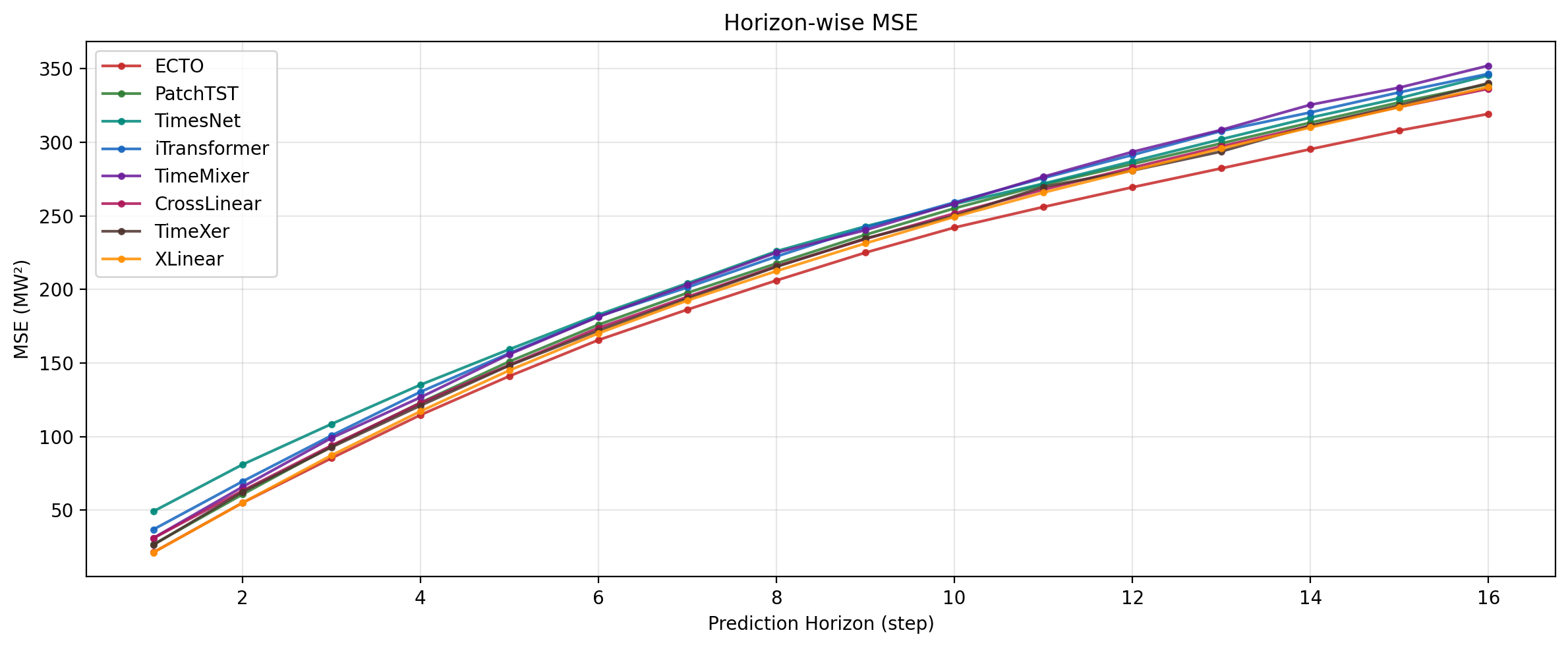}
\caption{Per-step MSE across the 16-step prediction horizon on WF1. ECTO maintains the lowest error at every horizon step.}
\label{fig:horizon-mse}
\end{figure}

\textbf{Consistent horizon-wise superiority.} Fig.~\ref{fig:horizon-mse} shows the per-step MSE across the 16-step prediction horizon. ECTO achieves the lowest MSE at every horizon step, with the margin over XLinear growing from 1.4\% at h1--4 (69.0 vs.\ 70.1) to 4.9\% at h13--16 (301.2 vs.\ 316.8). This indicates that ECTO's ECRR horizon refinement (Section~\ref{sec:ecrr}) provides increasingly beneficial calibration as the prediction horizon extends and the inherent uncertainty grows. Among the baselines, XLinear and CrossLinear are the closest competitors, while PatchTST, TimesNet, and iTransformer exhibit larger error growth at longer horizons.

\subsection{Interpretability Analysis}
\label{sec:interpret}

We analyze the learned behavior of PGVS and ECRR to verify that ECTO's exogenous conditioning mechanism produces physically meaningful and adaptive predictions. All analyses use the interpret model (Section~\ref{sec:setup}) with seed 2025. For cross-site comparability, the Xinjiang analysis uses the same 11 exogenous variable types as WF1 and WF4, excluding the redundant 70\,m wind-speed layer present in the full dataset (Table~\ref{tab:datasets}).

\subsubsection{PGVS: Adaptive Variable Selection}

\begin{figure}[htbp]
\centering
\includegraphics[width=\textwidth]{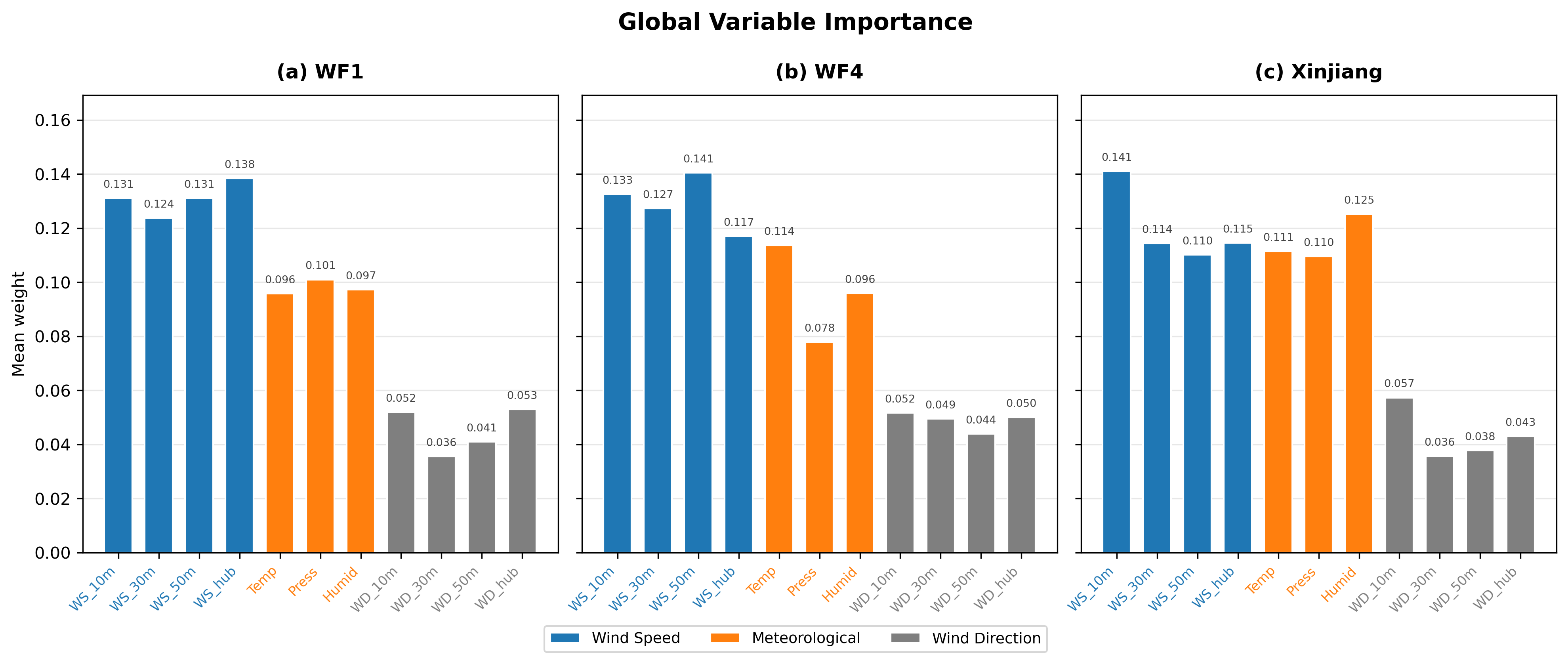}
\caption{Global average PGVS variable weights across three wind farms. Bars are color-coded by variable group: {\color{blue}wind speed}, {\color{orange}meteorological}, {\color{gray}wind direction}.}
\label{fig:var-importance}
\end{figure}

Fig.~\ref{fig:var-importance} shows the globally averaged PGVS variable weights for the three test wind farms. On all sites, wind-speed variables receive the highest weights overall (11--15\%), confirming that PGVS correctly identifies wind speed as the primary driver of short-term power output. Meteorological variables (temperature, pressure, humidity) contribute 11--12\%, while wind-direction variables contribute 3--6\%.

\begin{table}[htbp]
\centering
\caption{PGVS sparsity statistics across three wind farms. Active variables count those with weight $> 1/D$ where $D$ is the number of exogenous variables, chosen near the noise floor of the softmax selection output. Perplexity $= \exp(H)$ where $H = -\sum_i w_i \ln w_i$ is the entropy in nats.}
\label{tab:pgvs-sparsity}
\begin{tabular}{lccc}
\toprule
& WF1 & WF4 & Xinjiang \\
\midrule
Total exogenous variables & 11 & 11 & 11 \\
Mean active variables & 3.62 & 3.31 & 3.09 \\
Perplexity ($\exp(H)$) & 3.21 & 2.87 & 2.73 \\
Normalized entropy ($H / \ln D$) & 0.47 & 0.42 & 0.40 \\
\bottomrule
\end{tabular}
\end{table}

Table~\ref{tab:pgvs-sparsity} shows that PGVS activates only 3.1--3.6 variables per sample on average (out of 11 available), with a normalized entropy of 0.40--0.47. This indicates that the model consistently concentrates its attention on a small subset of variables while suppressing the rest, rather than distributing weight uniformly. The hierarchical group-then-variable structure (Section~\ref{sec:pgvs}) ensures that this sparsity operates within physically coherent groups, preventing the model from selecting an arbitrary combination of variables across groups. Across power regimes, PGVS maintains a stable sparse selection budget---the \emph{number} of active variables remains roughly constant, while \emph{which} variables are selected shifts with operating conditions.

\begin{figure}[htbp]
\centering
\includegraphics[width=\textwidth]{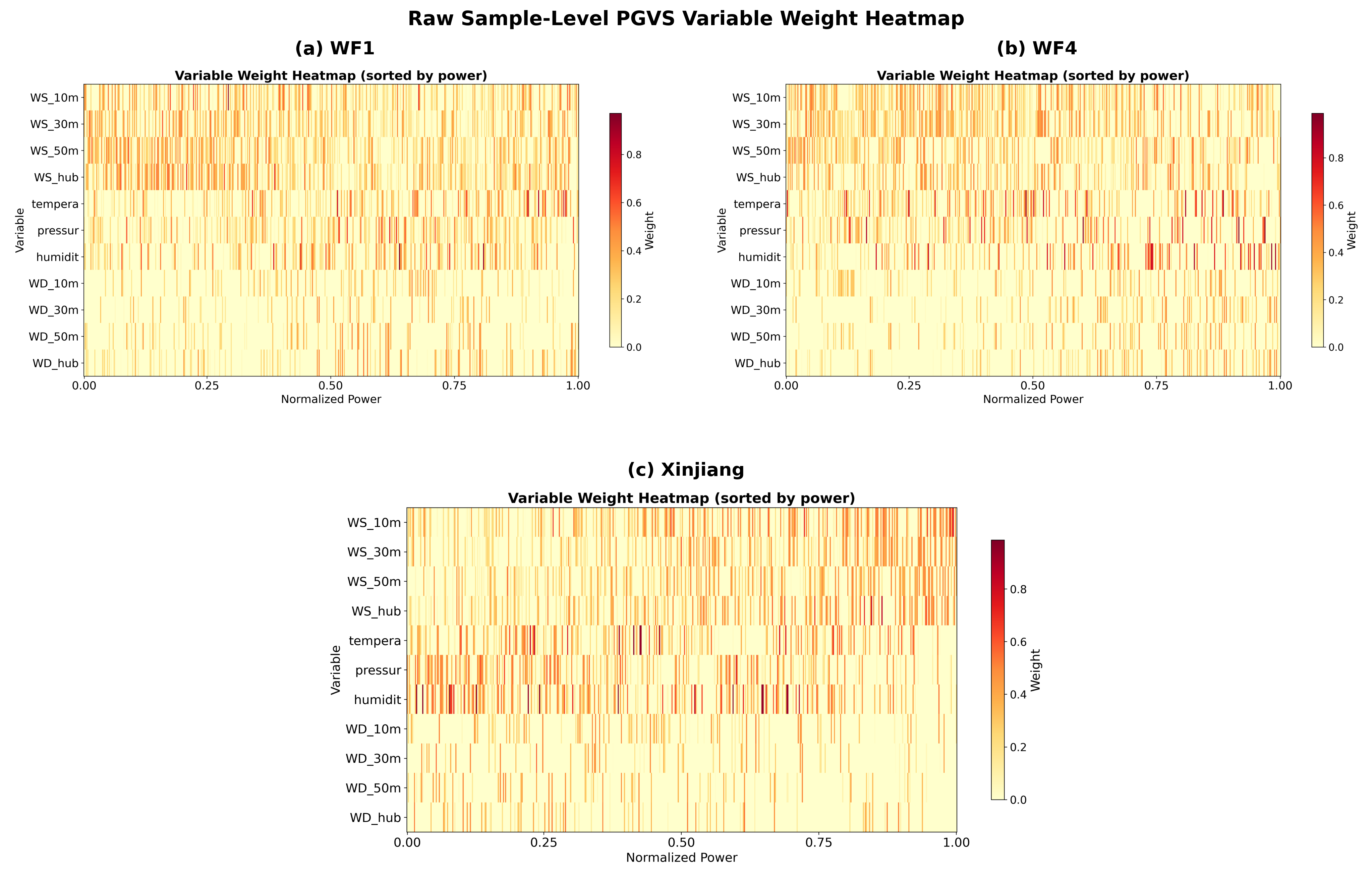}
\caption{Raw sample-level PGVS variable-weight heatmaps. Each panel shows individual test samples (sorted by normalized power, x-axis) versus exogenous variables (y-axis). The vertical stripe structure reflects sample-specific sparse activation: different samples activate different variable combinations depending on their operating conditions.}
\label{fig:pgvs-raw}
\end{figure}

To visualize this sample-level behavior, Fig.~\ref{fig:pgvs-raw} displays the raw PGVS weight maps for all three wind farms, with test samples sorted by normalized power. The prominent vertical stripe structure confirms that PGVS does not assign a fixed importance ranking---instead, it dynamically switches the active variable subset on a per-sample basis. The stripes are not random noise: within each site, the pattern of bright and dark bands shifts gradually along the power axis, suggesting that the activated variable combination correlates with operating conditions. However, the raw map also exhibits considerable sample-to-sample variability, making it difficult to extract systematic trends by visual inspection alone.

\begin{figure}[htbp]
\centering
\includegraphics[width=\textwidth]{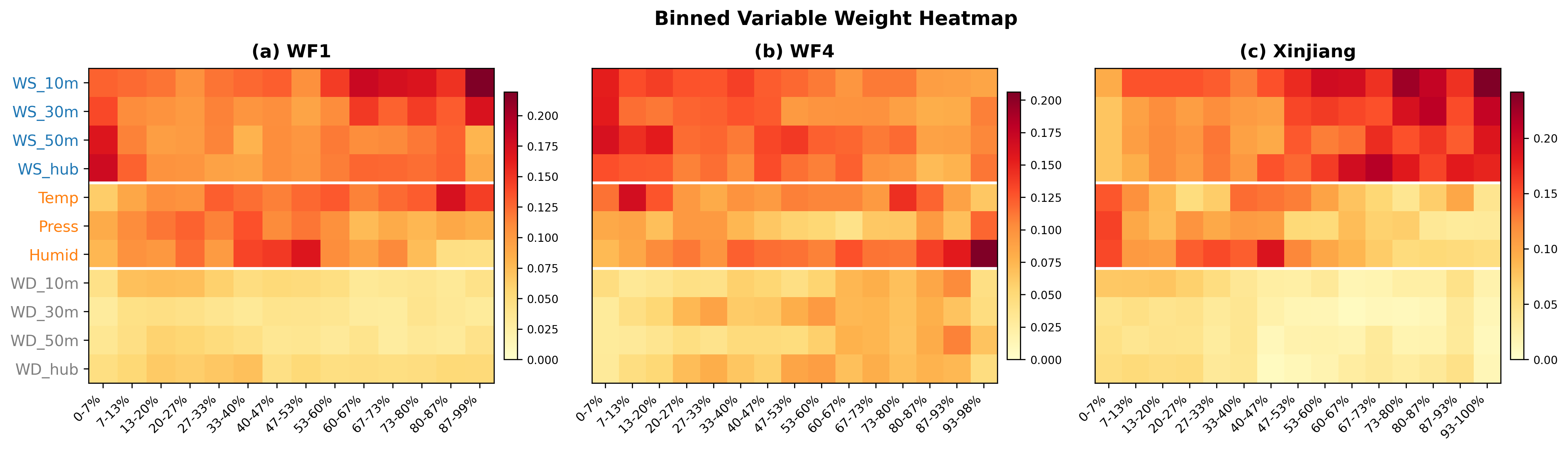}
\caption{PGVS variable-weight heatmaps averaged by power bin. Each row is an exogenous variable; color intensity indicates the mean PGVS weight within each bin. Variable names are color-coded by group: {\color{blue}wind speed}, {\color{orange}meteorological}, {\color{gray}wind direction}.}
\label{fig:pgvs-binned}
\end{figure}

Fig.~\ref{fig:pgvs-binned} aggregates the raw activations by power bin, resolving the sample-level variability into systematic operating-condition trends. The three wind farms exhibit qualitatively different adaptation patterns that reflect their distinct physical characteristics:

\textbf{WF1 (meteorological relay).} Wind-speed variables dominate at both low and high power levels, while temperature and pressure gain prominence in the mid-power range (20--50\%); wind-direction variables also show a transient mid-power peak (WD\_10m reaching 0.169 near 50\% capacity). This pattern is consistent with the physical picture that air density ($\rho = P / (R_d T)$) modulates power output most when the wind-speed contribution ($v^3$) is moderate, becoming less influential at very low and very high wind speeds.

\textbf{WF4 (humidity--wind-direction transition).} At high power levels, humidity becomes the dominant exogenous factor (rising from 0.072 to 0.206), while wind-speed weights decline slightly. Several wind-direction variables (WD\_30m, WD\_50m) also gain weight in the mid-to-high power range, which is consistent with wake effects where hub-height relative wind direction determines inter-turbine interference. The combined humidity--wind-direction pattern may reflect moist-air density modulation at higher wind speeds, amplified by the multi-turbine layout of the 66\,MW farm.

\textbf{Xinjiang (density-sensitive transition).} At this arid continental site with large diurnal temperature swings, temperature, pressure, and humidity collectively dominate at low power levels (with wind-direction variables maintaining modest weights of $\sim$0.07 before fading above 40\% capacity), while wind speed becomes the sole dominant factor above 60\% capacity. This is consistent with the power-curve physics: at low wind speeds, the air-density correction is comparable to the wind-speed contribution, whereas at high wind speeds the cubic relationship dominates.

These cross-site differences confirm that PGVS adapts its variable selection to local physical conditions rather than applying a fixed importance ranking. While wind-direction variables play a secondary role on WF1 and Xinjiang, their non-negligible weights indicate that PGVS retains directional information as a conditioning signal across all sites, with its relative importance determined by site-specific physical factors. Together with the sparse selection behavior quantified in Table~\ref{tab:pgvs-sparsity}, this demonstrates that PGVS provides a physically meaningful, condition-adaptive exogenous encoding, which directly conditions the regime router and calibration modules examined next.

\subsubsection{ECRR: Calibration Strategy Separation}

\begin{figure}[htbp]
\centering
\begin{subfigure}[b]{0.32\textwidth}
\centering
\includegraphics[width=\textwidth]{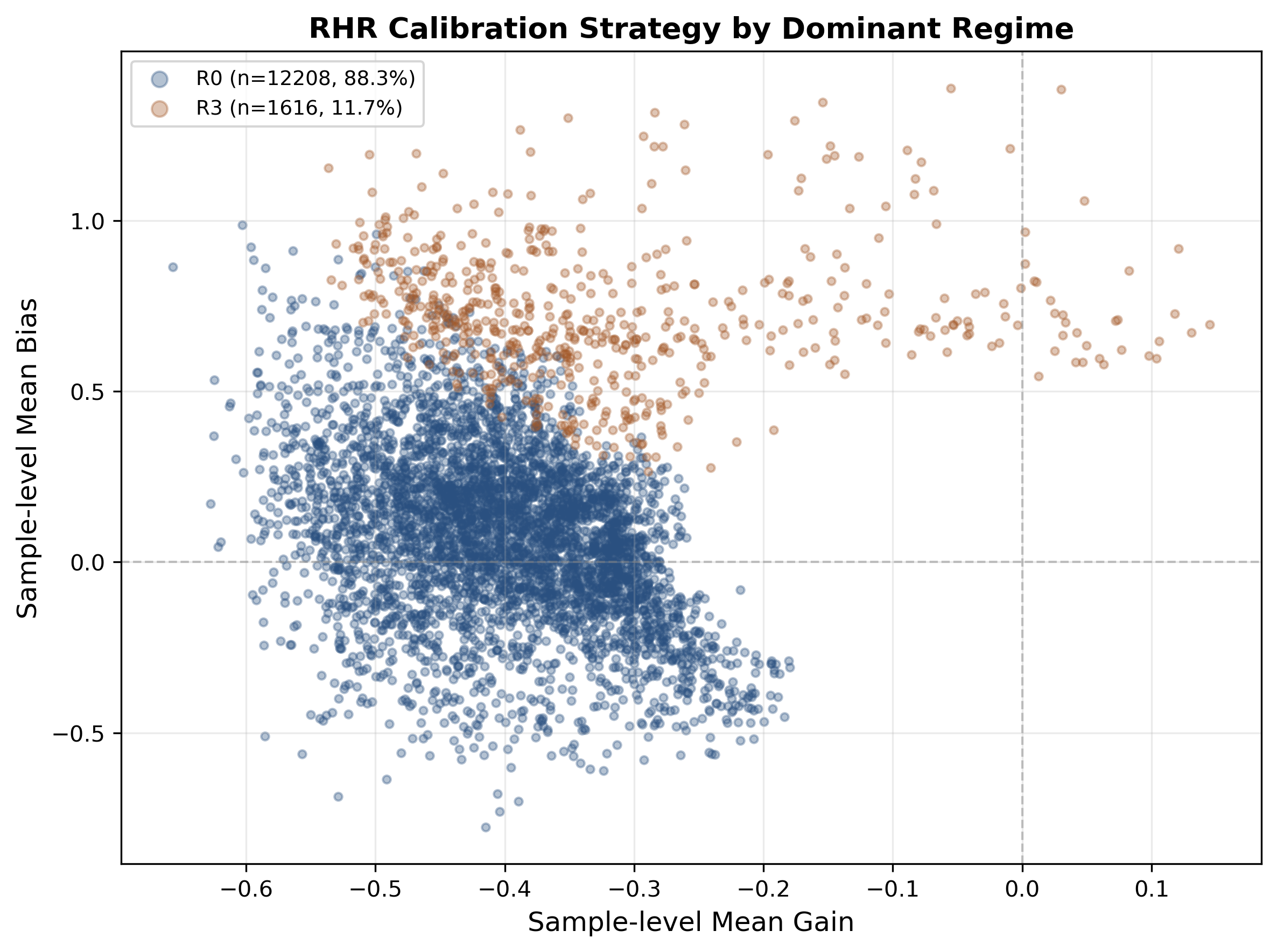}
\caption{WF1}
\label{fig:calib-wf1}
\end{subfigure}
\hfill
\begin{subfigure}[b]{0.32\textwidth}
\centering
\includegraphics[width=\textwidth]{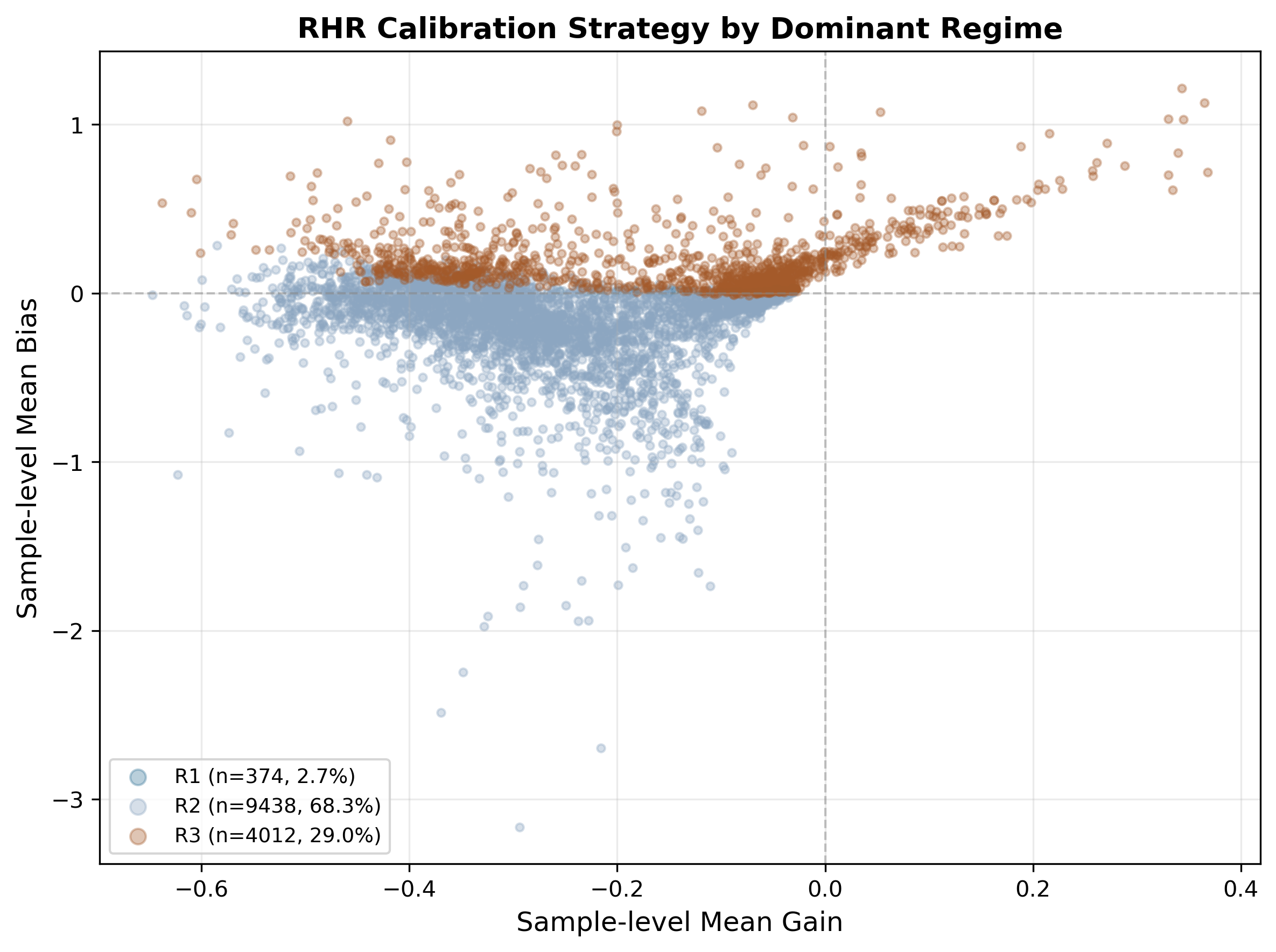}
\caption{WF4}
\label{fig:calib-wf4}
\end{subfigure}
\hfill
\begin{subfigure}[b]{0.32\textwidth}
\centering
\includegraphics[width=\textwidth]{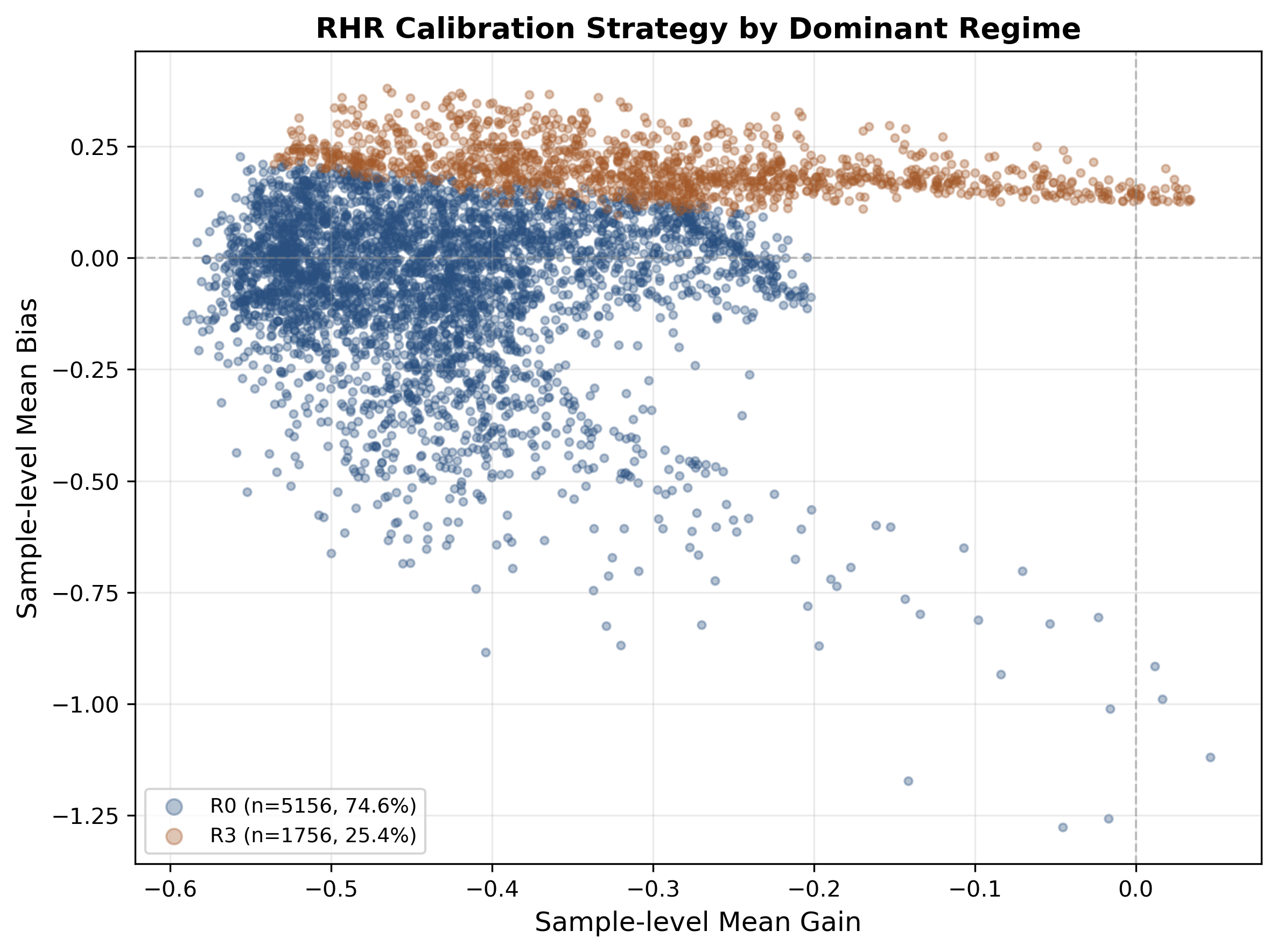}
\caption{Xinjiang}
\label{fig:calib-xinjiang}
\end{subfigure}
\caption{ECRR calibration strategy by dominant regime. Each point is a test sample; axes represent the sample-level mean gain and bias. Points are colored by the sample's dominant regime (the regime with the highest soft weight). The horizontal and vertical dashed lines mark zero gain and zero bias, respectively.}
\label{fig:calib-composite}
\end{figure}

The condition-adaptive exogenous context produced by PGVS is passed to ECRR, whose regime router maps it into a set of soft regime weights. These weights then blend multiple gain--bias calibration heads. Fig.~\ref{fig:calib-composite} visualizes this calibration behavior by plotting each test sample in the gain--bias plane, colored by its dominant regime.

Two structural properties of the learned calibration emerge from this visualization. First, the calibration points form \emph{two clearly separated clusters} across all three wind farms, indicating that the router has converged to two qualitatively distinct calibration strategies rather than a continuum or a single strategy. The dominant regime (R0 for WF1 and Xinjiang, R2 for WF4) applies a large negative gain with bias near zero, i.e., it \emph{scales down} the base prediction. The secondary regime (R3 across all sites) applies a smaller negative gain combined with a positive bias, i.e., it \emph{shifts up} the prediction. This emergent clustering into two strategies, rather than a continuum of adjustments, aligns with the design choice of a discrete regime router: the model learns to partition the exogenous condition space into a small number of distinct calibration behaviors, each appropriate for a different operating condition. We hypothesize that the two strategies target distinct sources of systematic error---scale-down likely addresses over-prediction, while shift-up corrects under-prediction---though their precise correspondence to specific error regimes awaits further analysis.

\begin{table}[htbp]
\centering
\caption{ECRR regime statistics across three wind farms. ``Dominant'' refers to the regime with the highest soft weight for each sample. Gain values are dimensionless multiplicative factors applied to the base forecast (RevIN-normalized space); bias values are additive offsets in the same normalized unit. All values are averaged over the 16 forecast horizons.}
\label{tab:regime-summary}
\begin{tabular}{lccc}
\toprule
& WF1 & WF4 & Xinjiang \\
\midrule
Primary regime & R0 & R2 & R0 \\
\quad Dominant \% & 88.3 & 68.3 & 74.6 \\
\quad Mean soft weight & 0.78 & 0.56 & 0.69 \\
\quad Gain / Bias & $-$0.40 / +0.08 & $-$0.27 / $-$0.19 & $-$0.43 / $-$0.06 \\
Secondary regime & R3 & R3 & R3 \\
\quad Dominant \% & 11.7 & 29.0 & 25.4 \\
\quad Mean soft weight & 0.14 & 0.25 & 0.27 \\
\quad Gain / Bias & $-$0.32 / +0.71 & $-$0.13 / +0.17 & $-$0.30 / +0.20 \\
\bottomrule
\end{tabular}
\end{table}

Note that the primary regime on WF4 (R2) behaves similarly to R0 on WF1 and Xinjiang---both apply a dominant negative gain---indicating that the router converges to the same two functional strategies across sites, with regime indexing being arbitrary.

Second, this two-strategy pattern is remarkably consistent across all three test sites despite their substantial differences in capacity (66--200\,MW), climate, and data distribution (Table~\ref{tab:datasets}). As summarized in Table~\ref{tab:regime-summary}, the secondary regime (R3) uniformly applies a more positive bias than the primary regime across all sites. While the qualitative two-strategy structure is preserved, the exact gain and bias magnitudes vary across sites---for instance, R3's bias ranges from +0.17 (WF4) to +0.71 (WF1)---reflecting the different power distributions and atmospheric characteristics of each wind farm. This site-specific calibration is a further indication that the router adapts to local conditions rather than learning a generic correction. The clean separation into two strategies---rather than a diffuse scatter---suggests that the PGVS-produced exogenous context carries sufficient information for the router to make structured calibration decisions, and that this architecture property generalizes beyond any single site.

Taken together, the PGVS and ECRR analyses show that ECTO's exogenous conditioning pipeline is end-to-end interpretable: PGVS performs sparse, condition-adaptive variable selection (Figs.~\ref{fig:var-importance}, \ref{fig:pgvs-raw} and~\ref{fig:pgvs-binned}, Table~\ref{tab:pgvs-sparsity}), and ECRR translates this selected context into a small set of well-separated calibration strategies (Fig.~\ref{fig:calib-composite}, Table~\ref{tab:regime-summary}).

\section{Conclusion}
\label{sec:conclusion}
\sloppy

This paper addressed the problem of ultra-short-term wind power forecasting from the perspective of structured exogenous variable modeling. Rather than treating meteorological inputs as generic auxiliary channels, we decomposed exogenous conditioning into two complementary questions---\emph{which} exogenous variables matter for the current prediction, and \emph{how} they should modulate the forecast under different operating conditions---and proposed the Exogenous-Conditioned Temporal Operator (ECTO) to answer both within a unified framework.

PGVS (Physically-Grounded Variable Selection) leverages a domain-informed\allowbreak physical grouping prior to perform hierarchical, sparse selection over candidate exogenous variables via sparsemax activations, producing a compact exogenous context that is conditioned on the target state and exogenous statistics of each sample. ECRR (Exogenous-Conditioned Regime Refinement) consumes this context to route the forecast through a small number of learned regime experts, each applying gain--bias calibration and horizon-specific corrections, without requiring pre-clustering or separate model training.

Experiments on three wind farms spanning different climates, capacities (66--200 MW), and exogenous dimensions (11--13 variables) demonstrate that ECTO achieves the lowest MSE across all sites. Relative improvements over the strongest baseline (XLinear) range from 2.2\% (WF4) to 5.2\% (Xinjiang), and the advantage is consistent across all 16 prediction horizons, extends to $H{=}32$, and is robust across random seeds. Ablation analysis confirms that each exogenous-related component contributes positively: removing all exogenous conditioning degrades MSE by 4.59\%, disabling PGVS by 1.84\%, and removing ECRR by 2.86\%. Interpretability analysis further reveals that PGVS learns site-specific variable selection patterns that align with known physical mechanisms---including air-density modulation, wind-direction-dependent wake effects, and density-sensitive transitions at arid continental sites---while ECRR converges to two well-separated calibration strategies (scale-down and shift-up) that are qualitatively consistent across all three test sites.

Several limitations warrant discussion. First, the evaluation is limited to single-site (turbine or farm-level) forecasting without modeling spatial dependencies such as inter-turbine wake effects; integrating spatial information through graph-based representations or NWP grid features is a natural extension. Second, the current framework produces point forecasts only; however, the ECRR module's mixture-of-experts structure naturally lends itself to probabilistic extensions---each expert could output distributional parameters (e.g., mean and variance), and the expert disagreement could serve as a measure of predictive uncertainty, which is increasingly important for reserve sizing and risk-aware dispatch. Finally, the proposed exogenous conditioning framework has only been validated on wind power; its applicability to other renewable energy domains with strong exogenous drivers, such as solar and hydro power forecasting, remains to be explored.

\section*{CRediT authorship contribution statement}
\textbf{Liyaqin Li}: Data curation, Investigation, Validation. \textbf{Junjun Wang}: Writing -- original draft, Visualization, Validation, Software, Methodology, Investigation. \textbf{Jianxiang Li}: Investigation, Validation. \textbf{Wei Huang}: Supervision, Resources. \textbf{Qianhui Yu}: Data curation. \textbf{Cao Yuan}: Writing -- review \& editing, Methodology, Conceptualization.

\section*{Data availability}
The WF1 and WF4 datasets are described by Chen and Xu~\cite{chen2022stategrid} and are publicly available via Figshare (DOI: 10.6084/m9.figshare.17304221.v4). The Xinjiang dataset is not publicly available due to confidentiality constraints.

\section*{Declaration of competing interests}
The authors declare that they have no known competing financial interests or personal relationships that could have appeared to influence the work reported in this paper.

\clearpage
\section*{Appendix. Supplementary material}

\begin{figure}[htbp]
\centering
\includegraphics[width=\textwidth]{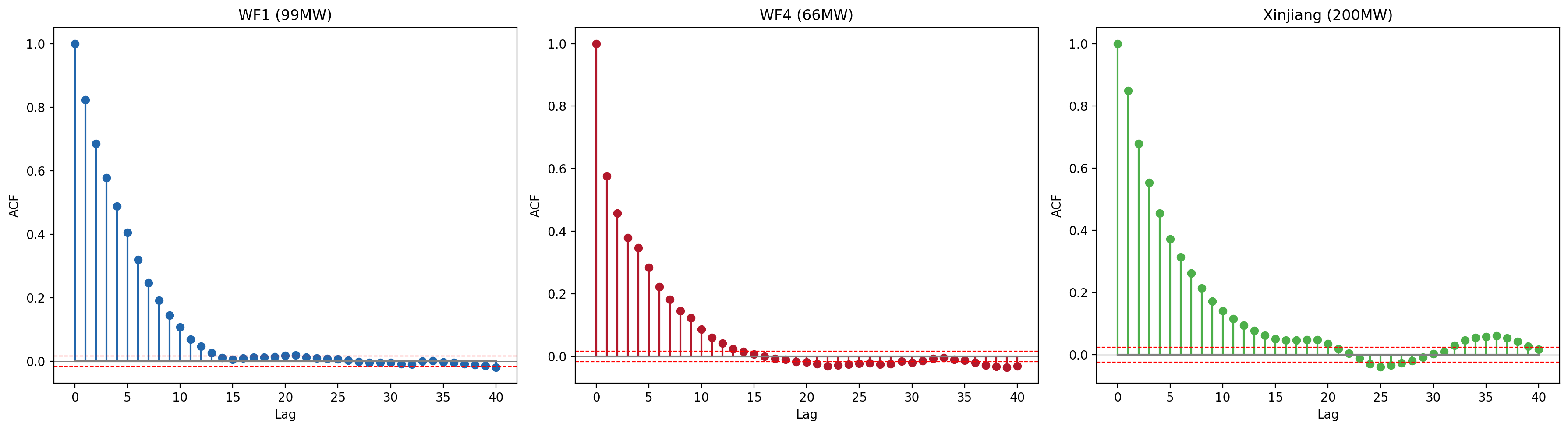}
\caption{Autocorrelation functions of the Diebold-Mariano loss differential series (ECTO vs.\ XLinear) for all three wind farms. The dashed red lines indicate the 95\% confidence interval under the white noise null. All three sites exhibit strong positive autocorrelation at short lags ($\rho_1 \approx 0.6$--$0.8$) that decays to near zero by $h = 16$, confirming that the Newey-West HAC bandwidth $h = H = 16$ (Bartlett kernel) used in Section~\ref{sec:main-results} adequately captures the persistent serial dependence in the forecast error differentials.}
\label{fig:dm-acf}
\end{figure}

\end{document}